\documentclass{article}

\PassOptionsToPackage{numbers, compress}{natbib}

\usepackage{multirow}
\usepackage{subcaption}
\usepackage{adjustbox}

\usepackage{array}      
\usepackage{colortbl}     
\usepackage{booktabs}     
\usepackage{caption}     
\usepackage{booktabs}
\usepackage{multirow}
\usepackage{amssymb}
\usepackage{pifont}
\usepackage{algpseudocode}
\usepackage{algorithm}
\usepackage{hyperref}       %
\usepackage[preprint]{neurips_2025}
\hypersetup{
    colorlinks=true,
    linkcolor=red,
    citecolor=blue,
    filecolor=magenta,      
    urlcolor=magenta,
}

\makeatother

\usepackage[utf8]{inputenc} %
\usepackage[T1]{fontenc}    %
\usepackage{hyperref}       %
\usepackage{url}            %
\usepackage{booktabs}       %
\usepackage{amsfonts}       %
\usepackage{nicefrac}       %
\usepackage{microtype}      %
\usepackage{xcolor}         %

\usepackage{array}
\usepackage{booktabs}
\usepackage{caption}
\usepackage{array} %
\usepackage{ragged2e} %
\usepackage[utf8]{inputenc}
\usepackage{amsthm}
\usepackage{algpseudocode}
\usepackage{tikz}
\usepackage{wrapfig}
\usepackage{amsmath}
\renewcommand{\paragraph}[1]{\vspace{-0.5mm}\noindent\textbf{#1}}

\title{REPA Works Until It Doesn’t: Early-Stopped, Holistic Alignment Supercharges Diffusion Training}

\author{
  Ziqiao Wang$^{1}$\footnotemark[1]\ \quad Wangbo Zhao$^{1}$\footnotemark[1]\ \quad Yuhao Zhou$^{1}$\ \quad Zekai Li$^{1}$\ \quad Zhiyuan Liang$^{1}$\ \\ \textbf{Mingjia Shi$^{1}$\ \quad Xuanlei Zhao$^{1}$\ \quad Pengfei Zhou$^{1}$\ \quad Kaipeng Zhang$^{2}$\footnotemark[2]\ } \\  \textbf{ Zhangyang Wang$^{3}$\ \quad Kai Wang$^{1}$\footnotemark[2]\ \quad Yang You$^{1}$ }
    \\[0.04cm]
    $^{1}$NUS HPC-AI Lab,
    $^{2}$Shanghai AI Laboratory,
    $^{3}$UT Austin
}
\begin{document}
\newcommand{\tablestyle}[2]{\setlength{\tabcolsep}{#1}\renewcommand{\arraystretch}{#2}\centering\footnotesize}
\renewcommand{\paragraph}[1]{\vspace{1.25mm}\noindent\textbf{#1}}
\newcommand\blfootnote[1]{\begingroup\renewcommand\thefootnote{}\footnote{#1}\addtocounter{footnote}{-1}\endgroup}

\newcolumntype{x}[1]{>{\centering\arraybackslash}p{#1pt}}
\newcolumntype{y}[1]{>{\raggedright\arraybackslash}p{#1pt}}
\newcolumntype{z}[1]{>{\raggedleft\arraybackslash}p{#1pt}}

\newcommand{\app}{\raise.17ex\hbox{$\scriptstyle\sim$}}
\newcommand{\mypm}[1]{\color{gray}{\tiny{$\pm$#1}}}
\newcommand{\x}{{\times}}
\definecolor{deemph}{gray}{0.6}
\newcommand{\gc}[1]{\textcolor{deemph}{#1}}
\definecolor{baselinecolor}{gray}{.9}
\newcommand{\baseline}[1]{\cellcolor{baselinecolor}{#1}}
\newcommand{\authorskip}{\hspace{2.5mm}}

\definecolor{dt}{HTML}{ADCAD8}
\definecolor{dt2}{HTML}{cddfe7}
\newcommand{\ots}[1]{\textcolor{dt}{#1}}
\newcommand{\otsmodel}[1]{\cellcolor{dt2}{#1}}

\definecolor{defaultcolor}{HTML}{E8E2F7}
\newcommand{\default}[1]{\cellcolor{defaultcolor}{#1}}

\let\cite\citep

\newcommand{\xmark}{\text{\ding{55}}}%
\newcommand{\dslope}{\text{\ding{216}}}%
\newcommand{\fslope}{\text{\ding{217}}}%
\newcommand{\uslope}{\text{\ding{218}}}%

\newcommand{\scolorbox}[2]{{\setlength{\fboxsep}{2pt}\colorbox{#1}{#2}}}

\renewcommand{\paragraph}[1]{\vspace{1.25mm}\noindent\textbf{#1}}
\newlength\savewidth\newcommand\shline{\noalign{\global\savewidth\arrayrulewidth
  \global\arrayrulewidth 1pt}\hline\noalign{\global\arrayrulewidth\savewidth}}
\newcolumntype{x}[1]{>{\centering\arraybackslash}p{#1pt}}
\newcolumntype{y}[1]{>{\raggedright\arraybackslash}p{#1pt}}
\newcolumntype{z}[1]{>{\raggedleft\arraybackslash}p{#1pt}}
\definecolor{degray}{gray}{.6}
\newcommand{\deemph}[1]{\textcolor{degray}{#1}}
\newcommand{\cmark}{\ding{51}}%

\maketitle
\renewcommand{\thefootnote}{\fnsymbol{footnote}}
\footnotetext[1]{equal contribution (ziqiaow@u.nus.edu). Ziqiao, Wangbo, Zhangyang, and Kai are core contributors.} \footnotetext[2]{corresponding author.}
\begin{abstract}

Diffusion Transformers (DiTs) deliver state-of-the-art image quality, yet their training remains notoriously slow.  
A recent remedy---\emph{representation alignment} (REPA) that matches DiT hidden features to those of a \emph{non-generative} teacher (e.g.\ DINO)---dramatically accelerates the \emph{early} epochs but plateaus or even degrades performance later.   We trace this failure to a \textbf{capacity mismatch}: once the generative student begins modelling the \emph{joint} data distribution, the teacher's lower-dimensional embeddings and attention patterns become a straitjacket rather than a guide. We then introduce \textbf{HASTE} (\textbf{H}olistic \textbf{A}lignment with \textbf{S}tage-wise \textbf{T}ermination for \textbf{E}fficient training), a two-phase schedule that keeps the help and drops the hindrance.  
Phase~\textit{I} applies a \emph{holistic} alignment loss that simultaneously distills \emph{attention maps} (relational priors) and \emph{feature projections} (semantic anchors) from the teacher into mid-level layers of the DiT, yielding rapid convergence. Phase~\textit{II} then performs one-shot termination that deactivates the alignment loss,
once a simple trigger such as a fixed iteration is hit, 
freeing the DiT to focus on denoising and exploit its generative capacity. HASTE speeds up training of diverse DiTs without architecture changes.  
On ImageNet $256{\times}256$, it reaches the vanilla SiT-XL/2 baseline FID in \textbf{50\,epochs} and matches REPA’s best FID in \textbf{500\,epochs}, amounting to a $\boldsymbol{28\times}$ reduction in optimization steps.  
HASTE also improves text-to-image DiTs on MS-COCO,
demonstrating to be a simple yet principled recipe for efficient diffusion training across various tasks.
Our {code} is available \href{https://github.com/NUS-HPC-AI-Lab/HASTE}{here}.
\end{abstract}

\section{Introduction}
\label{sec:intro}

\noindent
Diffusion Transformers (DiTs) are stunningly good—and stunningly slow. 
Recent variants such as DiT~\citep{Peebles2022DiT} and SiT~\citep{ma2024sitexploringflowdiffusionbased} achieve state‐of‐the‐art visual fidelity across a growing list of generative tasks~\citep{esser2024scalingrectifiedflowtransformers,flux2024,videoworldsimulators2024,feng2024dit4edit}.
Unfortunately, their training incurs vast compute and wall-clock budgets because each update must back-propagate through hundreds of noisy denoising steps.
A first wave of accelerators tackles this either by \emph{architectural surgery}—linearized attention, masking or gating~\citep{zheng2024fast,vavae,Hang_2023_ICCV,fasterdit,krause2025treadtokenroutingefficient}—or by \emph{training heuristics}, e.g.\ importance re-weighting of timesteps~\citep{wang2024closer}.
These interventions help, but often at the cost of specialized kernels or fragile hyper-parameter tuning.

\begin{wrapfigure}[15]{R}{0.45\textwidth}
    \centering
    \vspace{-1em}
    \includegraphics[width=0.45\textwidth]{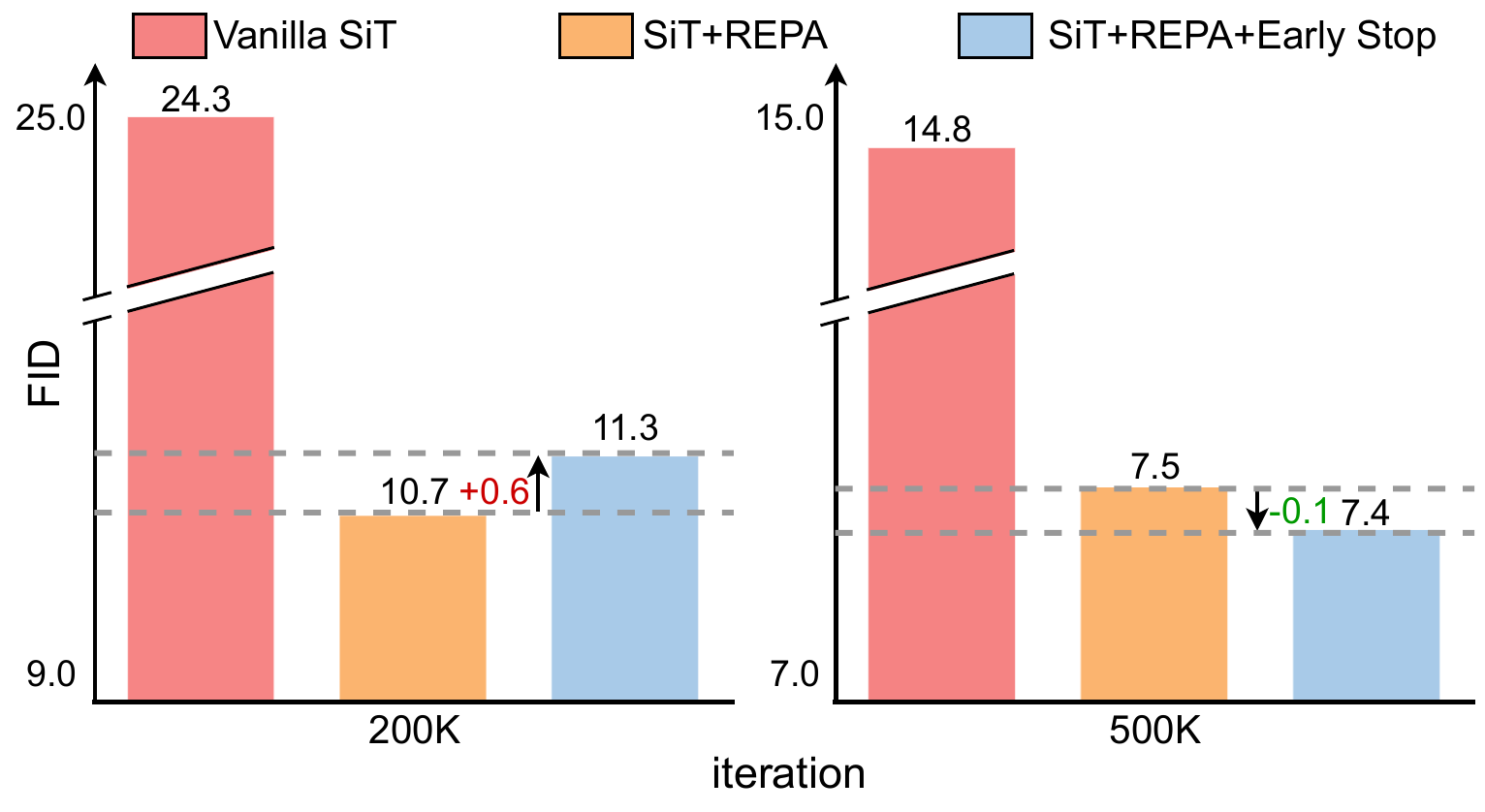}
    \caption{Training SiT-XL/2 on ImageNet $256{\times}256$.  
    Adding REPA slashes FID early on, but its benefit fades and ultimately reverses; dropping the alignment loss mid-training restores progress.}
        \vspace{-0.5em}
    \label{fig:XL-repa-es}
\end{wrapfigure}

\vspace{0.5em}
\paragraph{Representation alignment: early rocket, late parachute?} Recent work has demonstrated the effectiveness of leveraging external representations to accelerate diffusion model training—completely sidestepping the need for architectural modifications \citep{yu2025repa, vavae, tian2025urepaaligningdiffusionunets, leng2025repaeunlockingvaeendtoend}. A representative method, \emph{Representation Alignment} (REPA)~\citep{yu2025repa}, projects an intermediate DiT feature map onto the embedding space of a powerful \textbf{non-generative vision encoder} such as DINOv2~\citep{oquab2024dinov}, enforcing a cosine-similarity loss that bootstraps useful semantics during training.
The gain is immediate: the student DiT latches onto global object structure and converges several times faster than a vanilla run. Yet REPA’s help is not unconditional.
Figure~\ref{fig:XL-repa-es} removes the alignment loss after either $100\text{K}$ or $400\text{K}$ iterations.
Stopping \emph{late} (400K) \emph{improves} FID over the always-on baseline; stopping \emph{early} (100K) hurts—evidence that \textit{REPA works until it doesn’t}.
Why?

\vspace{0.5em}
\paragraph{Our Conjecture: Capacity mismatch incurrs the hidden turning point.}
Diffusion models eventually model the \emph{joint} data distribution, a harder objective than the \emph{marginal/conditional} targets implicit in a frozen, non-generative encoder.
Consequently, once the student has burned in, its own capacity overtakes the teacher’s.
Our gradient-angle analysis (Section~\ref{sec:es}) shows alignment and denoising objectives start \emph{aligned} (acute angles), drift to orthogonality, then turn obtuse—signalling that continued alignment may become a harmful constraint.

\vspace{0.5em}
\paragraph{Simple Remedy: Holistic alignment, then release.}
Two observations motivate our remedy.
First, the teacher’s \emph{attention maps} encode relational priors that are as valuable as its embeddings~\citep{li2024on,hoang2024compressedllmsforgetknowledge}; guiding only features leaves this structural knowledge untapped.
Second, the alignment needs a \emph{stage-wise schedule}: thick guidance early, zero guidance once gradients diverge. We therefore introduce \textbf{HASTE} (\textbf{H}olistic \textbf{A}lignment with \textbf{S}tage-wise \textbf{T}ermination for \textbf{E}fficient training).
During \underline{Phase~I} we distill \emph{both} projected features \emph{and} mid-layer attention maps from DINOv2 into the DiT, giving the student relational and semantic shortcuts.
Once a simple trigger (\emph{e.g.}, fixed iteration or gradient-angle threshold) is hit, we enter \underline{Phase~II}: the alignment loss is disabled and training proceeds with the vanilla denoising objective.
The recipe is two lines of code, no kernel changes.

\vspace{0.5em}
\paragraph{Contribution Summary.} Our findings refine the community’s understanding of external representation guidance: it is immensely helpful early, but \emph{must be let go} for the generative model 
to focus on specific tasks.
We outline our contributions as follows.
\begin{itemize}
    \item \textbf{Diagnosis.}  We identify a capacity mismatch that flips REPA from accelerator to brake and quantify it via gradient-direction similarity.
    \item \textbf{Method.}  We propose \emph{holistic} (attention\,+\,feature) alignment combined with a \emph{stage-wise termination} switch that deactivates alignment when it starts to impede learning.
    \item \textbf{Results.}  On ImageNet $256{\times}256$ our schedule matches vanilla SiT-XL/2 in \textbf{50\,epochs}, amounting to a \textbf{$28\times$} speed-up, and reaches REPA’s best score in \textbf{500\,epochs}.  Gains replicate on COCO text-to-image generation task.
\end{itemize}

\begin{wrapfigure}{T}{0.46\textwidth}
\vspace{-6em}
  \centering
  \includegraphics[width=0.46\textwidth]{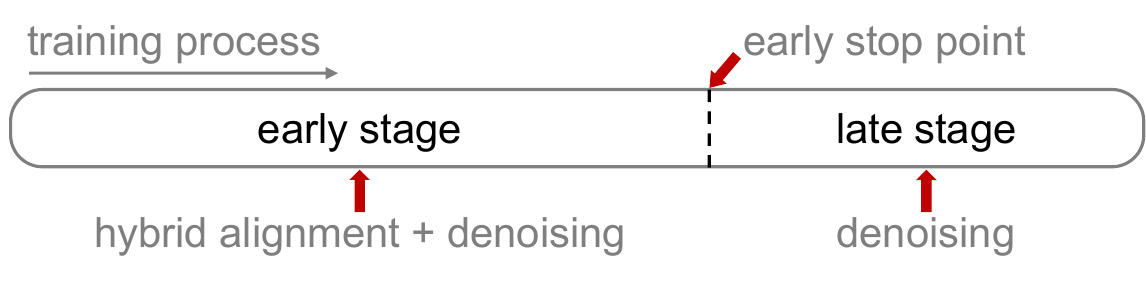}
\caption{Overview of our framework. \emph{Phase I} (left) distills \textit{both} feature embeddings and attention maps from a frozen, non-generative teacher (DINOv2) into mid-level layers of the student DiT.   When a simple trigger $\tau$ fires, the alignment loss is \emph{disabled}; \emph{Phase II} (right) then continues training with pure denoising.}
\vspace{-1em}
\label{fig:RAES-Pipeline}
\end{wrapfigure}

\section{Method}\label{sec:Method}

Our framework, HASTE, couples two ingredients (see Figure~\ref{fig:RAES-Pipeline}: (i) \textbf{Holistic alignment}: a \emph{dual‐channel} distillation that supervises both projected features and attention maps; (ii) \textbf{Stage-wise termination}: a \emph{single switch} that turns the alignment loss off once it ceases to help. We first recap REPA and attention alignment, then describe how we marry them and when we shut them off.

\subsection{Preliminaries}
\label{sec:prelims}

\paragraph{Notation.}
Let $\mathbf{x}$ be a clean image, $\tilde{\mathbf{x}}_t$ its noised version at timestep~$t$, and $\mathbf{h}_t$ the hidden state of a Diffusion Transformer $\mathcal{G}_\theta$.
A frozen, non-generative vision encoder $\mathcal{E}$ (DINOv2) produces patch embeddings $\mathbf{y}=\mathcal{E}(\mathbf{x})$ and self-attention matrices $\mathbf{A}^E$.

\paragraph{Representation alignment (REPA).}
A small MLP~$g_\phi$ projects $\mathbf{h}_t$ into the encoder space. REPA \cite{yu2025repa} then aligns the projected state $g_{\phi}(h_t)$ with $y$ by maximizing token-wise cosine similarities:
\begin{equation}
  \mathcal{L}_{\text{REPA}}(\theta,\phi)=
  -\mathbb{E}_{\mathbf{x},\epsilon,t}\left[\frac{1}{N} \sum_{n=1}^{N} \mathrm{sim}\left(\mathbf{y}^{[n]}, g_\phi\left(\mathbf{h}_{t}^{[n]}\right)\right)\right]
  \label{eq:repa}
\end{equation}
This regularization is jointly optimized with the original denoising objective,  to guide the more efficient training of diffusion transformers.

\paragraph{Attention alignment (ATTA).} 
 ATTA aims to transfer attention patterns from a pre-trained teacher model to a student model to guide the latter’s training process \citep{li2024on}. For selected layers/heads $(i,j)$ we minimize token-wise cross-entropy between teacher and student attention.

\subsection{Early Stop of Representation Alignment}
\label{sec:es}

\paragraph{Gradient–based autopsy reveals state evolution.}
Figure~\ref{fig:XL-repa-es} already hinted that REPA’s benefit peaks early and tapers off.  
To pinpoint \emph{when} the auxiliary loss flips from help to hindrance, we inspect the \emph{cosine similarity}  
\[
    \rho_t
    \;=\;
    \cos\!\bigl(
        \nabla_\theta \mathcal{L}_{\text{diff}},\, 
        \nabla_\theta \mathcal{L}_{\text{REPA}}
    \bigr)
    \quad\in[-1,1],
\]
computed on the $8^{\text{th}}$ block of SiT--XL/2 (the alignment depth used by REPA) over \(960\) ImageNet images (see details in Appendix~\ref{app:grad_ang}).
A positive $\rho_t$ means the teacher pushes the student in roughly the \emph{same} direction as denoising; negative means the two losses actively fight.

\begin{wrapfigure}[8]{r}{0.5\textwidth}
    \vspace{-3em}
    \centering
    \includegraphics[width=0.5\textwidth]{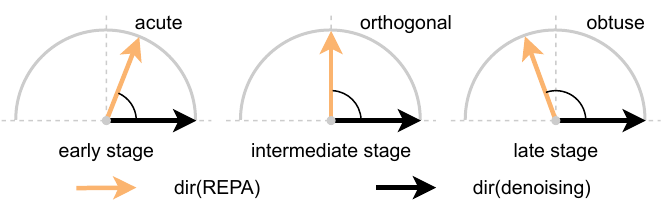}
    \caption{{Cosine similarity between REPA and denoising gradients.  
    Acute $\to$ orthogonal $\to$ obtuse: the auxiliary signal turns from booster to brake.}}
    \label{fig:grad_sim_stage}
    \vspace{-3em}
\end{wrapfigure}

Taking $t\leq0.1$ for example, Figure~\ref{fig:grad_sim_stage} shows three distinct regimes:

\begin{enumerate}
    \item \emph{Ignition} (0–200 K iters): 
    $\rho_t$ starts with a relatively high level --- REPA \textbf{adds} power; diffusion transformer profits from the teacher’s guidance on representation learning.
    \item \emph{Plateau} (200 K–400 K iters): 
    $\rho_t$ decreases to nearly orthogonal level --- objectives decouple; further REPA updates neither help nor hurt.
    \item \emph{Conflict} ($>\!400$ K iters): 
    $\rho_t$ exhibits negative values --- gradients oppose; REPA now \textbf{erases} detail the student tries to learn.
\end{enumerate}

The cross–over coincides with the iteration where Figure~\ref{fig:XL-repa-es} shows FID curves diverging, confirming that gradient geometry is a faithful early-warning signal.

\begin{wrapfigure}[8]{r}{0.5\textwidth}
\vspace{-3em}
    \centering
    \includegraphics[width=0.5\textwidth]{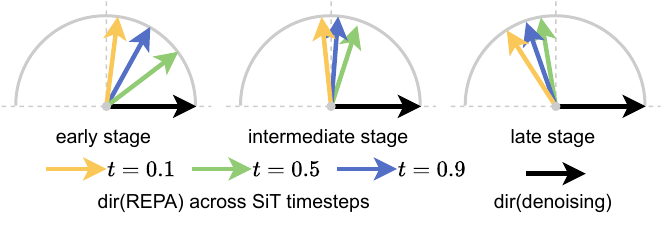}
    \vspace{-1.5em}
    \caption{{Gradient similarity as function of diffusion timestep \(t\).  
    At \(t\!=\!0.1\) (high-detail phase) the two losses already conflict even early in training.}}
\vspace{-3em}
    \label{fig:grad_sim_time}
\end{wrapfigure} 

\paragraph{Why does conflict arise?  Capacity–mismatch view.}
Once the student starts modelling the \emph{joint} data distribution, it seeks high-frequency detail absent from the teacher’s embeddings.
A frozen encoder trained for invariant recognition discards such minutiae by design; forcing the student back into that lower-dimensional manifold yields destructive gradients.  
We see the same mismatch at the level of \emph{diffusion timesteps.}

Figure~\ref{fig:grad_sim_time} plots $\rho_t$ versus the diffusion time index.
For mid-noise steps (e.g., $t=0.5$) where the image is still blurry, gradients align.
For late steps ($t\!\le\!0.1$)---responsible for textures and fine grain~\citep{DBLP:conf/icml/KimSSKM22}---they are near-orthogonal \emph{from the start}.
This indicates that teacher guidance is intrinsically global; when the denoiser must polish pixels, the encoder has little to teach.

\begin{wrapfigure}[14]{r}{0.52\textwidth}
\vspace{-1.5em}
    \centering
    \includegraphics[width=0.50\textwidth]{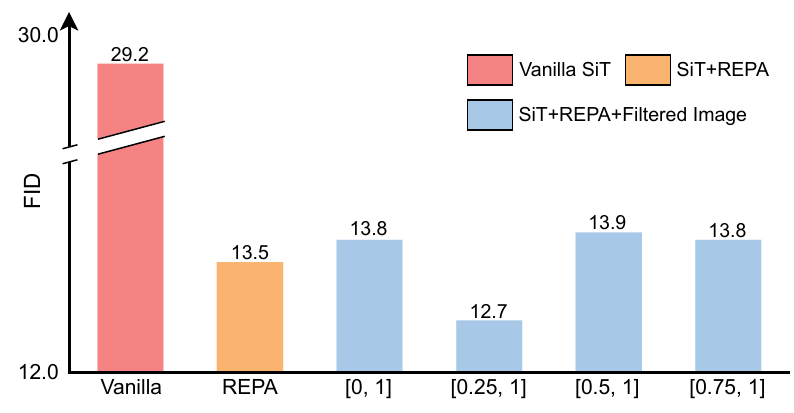}
    \caption{{Replacing teacher inputs with low-pass images leaves REPA’s early gain intact: evidence that the auxiliary loss transmits mainly global structure. We train SiT-L/2 for 200K iterations.}}
    \label{fig:repa_intervals}
\end{wrapfigure}

We sharpen this claim by feeding the teacher \emph{low-frequency only} versions of each image (Figure~\ref{fig:repa_intervals}).
Early FID improves almost identically to vanilla REPA, proving that the speed-up stems from \emph{coarse semantic scaffolding}; high-frequency cues are irrelevant to REPA’s benefit.

\paragraph{Take‐away.}
REPA supplies valuable \emph{global} context but obstructs \emph{local} detail once the student matures.
Hence alignment should be \textbf{transient} for further improvement.

\vspace{0.3em}
\noindent\textbf{Fix: Stage-wise termination.}  
Let $\tau$ denote the termination iteration around which $\rho_t$ exhibits low similarity and the alignment provides limited benefit.
We then \emph{discard} the auxiliary alignment loss:
\begin{equation}
\label{eq:earlystop}
  \mathcal{L}(\theta,\phi)=
  \begin{cases}
     \mathcal{L}_{\text{diff}} + \mathcal{L}_{R}, & n<\tau,\\[2pt]
     \mathcal{L}_{\text{diff}}, & n\ge\tau,
  \end{cases}
\end{equation}
where $\mathcal{L}_{R}$ may itself be the holistic combo of feature {(Section~\ref{sec:prelims})} and attention {Section~\ref{sec:attn})} losses.
A fixed $\tau$ works nearly as well but the gradient rule adds robustness across datasets.

\begin{figure*}[htbp]
  \centering
     \begin{subfigure}{0.46\linewidth}
        \includegraphics[height=0.53\linewidth]{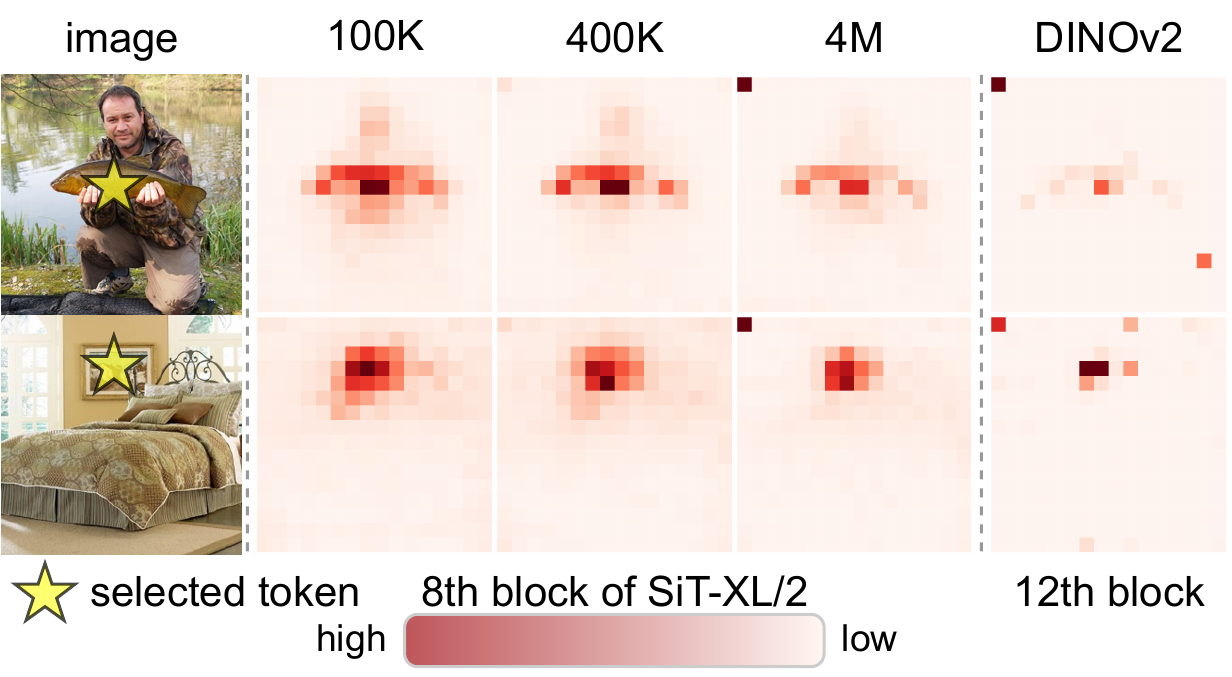}
        \caption{Visualization of attention maps from DINOv2-B and SiT-XL/2+REPA at different training iterations.}
        \label{fig:repa_attn_maps}  
     \end{subfigure}
     \hfill
      \begin{subfigure}{0.26\linewidth}
        \includegraphics[height=0.89\linewidth]{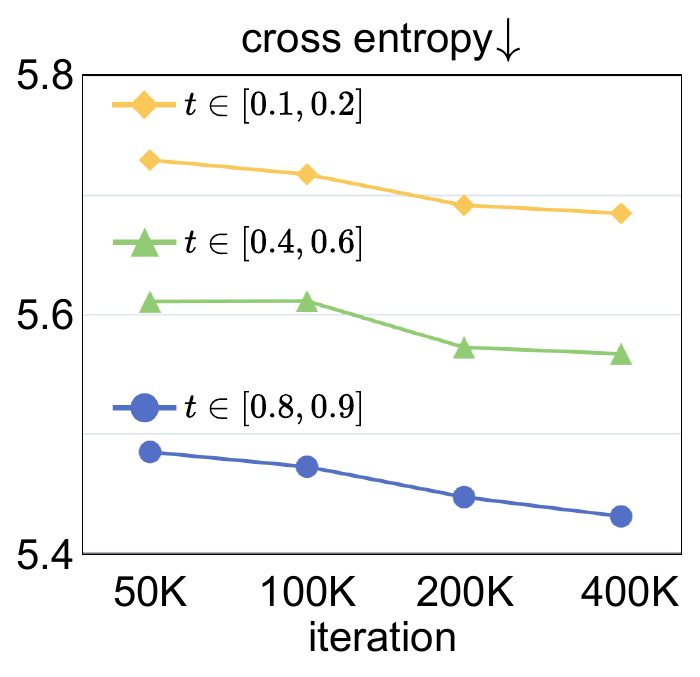}
        \caption{Attention alignment progress with REPA alone.}
        \label{fig:repa-test-attn}
      \end{subfigure}
      \hfill
      \begin{subfigure}{0.26\linewidth}
        \includegraphics[height=0.89\linewidth]{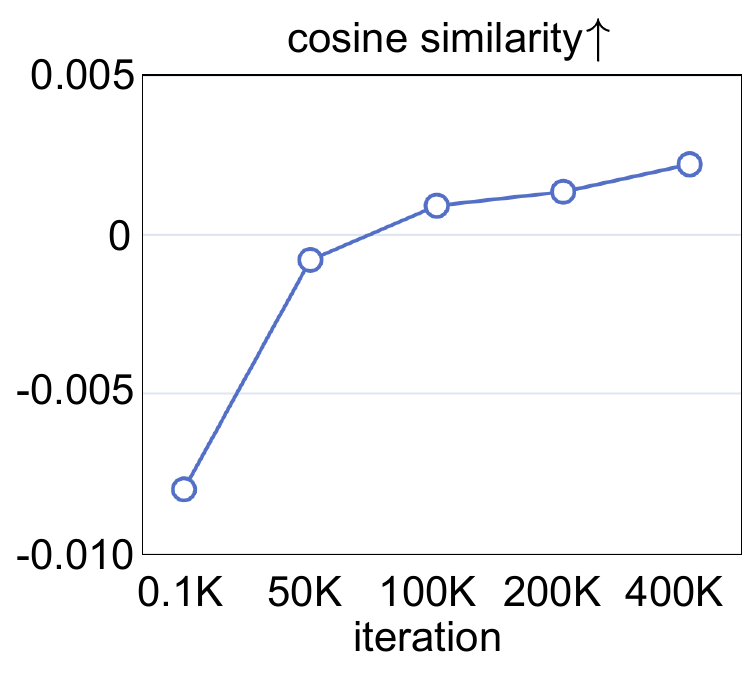}
        \caption{Feature alignment progress with ATTA alone.}
        \label{fig:attn-test-repa}
      \end{subfigure}
      \caption{Evaluating cross-effects between feature and attention alignment.
(a) Attention map visualization of selected tokens for SiT-XL/2+REPA and DINOv2-B.
(b) Alignment depth at 5, we track attention map cross-entropy between the 12th-layer of DINOv2-B and the 5th-layer of SiT-B/2.
(c) Attention maps from 3rd–5th layers of SiT-B/2 are aligned with those from 8th, 10th, and 12th layer of DINOv2-B. Since ATTA alone does not optimize the projector, we directly compute cosine similarity between the DINOv2-B features and the 5th-layer hidden states of SiT (without projection).}
      \label{fig:3.4}
\end{figure*}

\subsection{Holistic Alignment by Integration Attention}
\label{sec:attn}

\paragraph{Rationale: Why attending to \emph{attention}?} Compared with \emph{token embeddings}, self–attention matrices reveal \emph{where} a transformer routes information at each layer—its ``inference pathways’’ in the sense of \citet{hoang2024compressedllmsforgetknowledge}.  
These pathways encode rich relational priors: object–part grouping, long–range symmetry, and background–foreground segregation emerge as distinct heads in DINOv2, even though the model was trained without labels.  
Critically, such routing information is \emph{orthogonal} to the static content captured by features: two models can share identical patch embeddings but attend to them in entirely different patterns, leading to divergent downstream behavior.

Recent evidence echos: \citet{li2024on} show that distilling \emph{only} attention maps from a high-capacity teacher to a randomly initialized ViT is more effective than transferring \emph{only} embeddings, in recovering the teacher's linear probe precision on ImageNet.   
The asymmetry suggests that attention acts as a \textbf{structural prior}: once the model is taught \emph{how to look}, it can relearn \emph{what to look at} rapidly.

For diffusion transformers, they must integrate global spatial cues (layout, object boundaries) across hundreds of tokens for effective representation construction.
While feature alignment (REPA) accelerates the learning process by injecting semantic anchors, the structural knowledge remains underexploited.
Attention alignment targets the complementary regime: it transfers the \emph{global routing template} to DiT, thereby enabling precise spatial and global information guidance.

\paragraph{Motivational experiments.} To disentangle the respective contributions of \textit{features} and \textit{attention}, we probe the two signals in isolation:
\begin{enumerate}
    \item[\textit{(i)}] \textbf{Feature alignment only (REPA).}  
      Figure~\ref{fig:repa_attn_maps} shows that REPA gradually makes SiT heads resemble those of the teacher. However, the convergence of attention patterns is slow and incomplete (see cross-entropy trend in Figure~\ref{fig:repa-test-attn}).
    \item[\textit{(ii)}] \textbf{Attention alignment only (ATTA).}  
     Aligning attention maps alone can also pull the student’s hidden features toward the teacher’s embedding space (Figure~\ref{fig:attn-test-repa}) and yields a training-speed boost on par with REPA (see details in Section~\ref{AblationStudies}).
\end{enumerate}
\textbf{Takeaway.} REPA bootstraps \emph{semantics} but leaves routing under-constrained; ATTA nails routing but still requires the conditional gates to be learned from scratch.  
Their complementary effects motivate combining both.
For a chosen set $\mathcal{S}$ of student–teacher layer pairs $(\ell_s,\ell_t)$ and the $M$ heads,  
\begin{equation}
\label{eq:atta-loss}
  \mathcal{L}_{\text{ATTA}}
  \;=\;
  \frac{1}{|\mathcal{S}|\,M}
  \sum_{(\ell_s,\ell_t)\in\mathcal{S}}
  \sum_{m=1}^{M}
  \mathcal{H}\!\Bigl(
     \text{softmax}\!\bigl(\!Q^{\ell_s,m}_s K^{\ell_s,m\top}_s\bigr),
     \text{softmax}\!\bigl(\!Q^{\ell_t,m}_t K^{\ell_t,m\top}_t\bigr)
  \Bigr),
\end{equation}
where $\mathcal{H}$ is token–wise cross-entropy.

\paragraph{Where and when to align attention?} We distill teacher heads \emph{only} into \textbf{intermediate} student blocks (e.g., SiT-XL/2 blocks 4–7).  
Two empirical observations justify this selective schedule:

\begin{enumerate}
\item[\textit{(i)}] \textbf{Shallow mismatch.}  
Early DiT layers ingest \emph{Gaussian-noisy latents}; their representations are dominated by variance normalization and channel folding rather than semantics.  
Supervising those layers with \emph{pixel-space} attention from a clean-image encoder is therefore off-manifold.  
In practice, forcing attention on too many shallow layers destabilizes the loss and raises FID.

\item[\textit{(ii)}] \textbf{Deep freedom.}  
The ultimate objective of DiT is denoising for high-quality generation, rather than representation learning. The last blocks are responsible for translating high-level structure into precise generation update.
Thus, these blocks should remain dedicated to the denoising objective, unregularized.
\end{enumerate}

Aligning mid-layers strikes the sweet spot:
they are late enough that latents carry discernible semantics, yet early enough that constraining their routing gives downstream blocks a clean, well-organized feature tensor to refine.

\subsection{Final Recipe: HASTE}
\label{sec:integ}

\textbf{Where we align.}  
\emph{Attention maps} from the teacher are distilled into a \textit{range} of mid-depth DiT blocks;  
\emph{features} follow the original REPA setting—one projection at a single mid-layer.  
Neither the shallow noise processing blocks nor the final denoising blocks are regularized.

\textbf{What we align.}  
During Phase I (iterations $n<\tau$) we apply a \emph{hybrid} auxiliary loss. 
$\lambda_{R}$ and $\lambda_{A}$ are weight coefficients for balancing two regularizations. 
\begin{equation}
\label{eq:hybrid}
  \mathcal{L}_{R}
  = \lambda_{R}\,\mathcal{L}_{\mathrm{REPA}}
  + \lambda_{A}\,\mathcal{L}_{\mathrm{ATTA}}.
\end{equation}

\textbf{When we stop.}  
At the switch point $\tau$—chosen as a fixed iteration or the gradient-angle trigger from Section~\ref{sec:es}—\emph{both} terms in \eqref{eq:hybrid} are dropped and training proceeds with the vanilla denoising objective.

This three-line schedule constitutes \emph{HASTE}: it retains REPA’s semantic anchoring, adds ATTA’s routing prior, and removes all auxiliary constraints once they turn counter-productive.

\section{Experiments}
\label{sec:exp}
\subsection{Setup}
\paragraph{Models and datasets.} 
Following REPA \citep{yu2025repa}, we conduct experiments on three diffusion transformers: SiT \citep{ma2024sitexploringflowdiffusionbased}, DiT \citep{Peebles2022DiT}, and MM-DiT \citep{esser2024scalingrectifiedflowtransformers}. ImageNet \citep{ImageNet} and MS-COCO 2014 \citep{MSCOCO} datasets are used for class-to-image and text-to-image generation tasks, respectively. Moreover, we employ a pre-trained DINOv2-B \citep{oquab2024dinov} as the representation model to extract high-quality features and attention patterns.

\paragraph{Implementation details.} 
We use a training batch size of 256 and SD-VAE \citep{LDM} for latent diffusion, and set $\lambda_{R} = 0.5$ following REPA to ensure a fair comparison.  
Additionally, we also adopt the SDE Euler-Maruyama sampler with $\text{NFEs} = 250$ for image generation on SiT and DiT. 
We set $\lambda_A=0.5$ as the weight of attention alignment. We use NVIDIA A100 and H100 compute workers.

\paragraph{Evaluation metrics.} For ImageNet experiments, we sample 50K images to assess the performance, leveraging evaluation protocols provided by ADM \citep{ADM} to measure FID \citep{10.5555/3295222.3295408}, sFID \citep{pmlr-v139-nash21a}, IS \citep{IS}, and Precision and Recall \citep{precandrec}. For text-to-image generation, we follow the settings defined in \citep{bao2022all}. 

\subsection{Experiments on ImageNet 256 $\times$ 256}

\paragraph{Setting.}
In this experiment, we set the termination point $\tau = 100$K iteration (around 20 epochs) for SiT-B/2 and $\tau = 250$K iteration (around 50 epochs) for large and xlarge size models.
while all other settings remain at their default values.

\begin{wraptable}[32]{r}{0.55\textwidth}
\tablestyle{2.7pt}{1.3}
\vspace{-1em}
\centering
\begin{tabular}{lcccccc}
method                             & epoch & FID$\downarrow$ & sFID$\downarrow$ & IS$\uparrow$   & Prec.$\uparrow$ & Rec.$\uparrow$ \\ \midrule[1.2pt]
\multicolumn{7}{c}{\emph{Without} Classifier-free Guidance (CFG)} \\
MaskDiT                            & 1600   & 5.69             & 10.34            & \textbf{177.9} & \textbf{0.74}   & 0.60           \\
DiT               & 1400 & 9.62          & 6.85          & 121.5 & 0.67 & 0.67       \\
SiT               & 1400 & 8.61          & 6.32          & 131.7 & 0.68 & 0.67       \\
DiT+REPA        & 170  & 9.60          & -             & -     & -    & -          \\
SiT+REPA        & 800  & 5.90          & 5.73          & 157.8 & 0.70 & \textbf{0.69} \\
FasterDiT         & 400  & 7.91          & 5.45          & 131.3 & 0.67 & \textbf{0.69} \\
MDT               & 1300 & 6.23          & 5.23          & 143.0 & 0.71 & 0.65       \\
\rowcolor[HTML]{efefef}  
DiT+\textbf{HASTE}        & 80   & 9.33 & 5.74  & 114.3          & 0.69          & 0.64       \\
\rowcolor{gray!8} 
\cellcolor{gray!8} & 50   & 8.39 & 4.90  & 119.6          & 0.70          & 0.65       \\
\rowcolor{gray!8} 
\multirow{-2}{*}{\cellcolor{gray!8}SiT+\textbf{HASTE}} & 100    & \textbf{5.31}    & \textbf{4.72}    & 148.5        & 0.73            & 0.65

\\\hline
\multicolumn{7}{c}{\emph{With} Classifier-free Guidance (CFG)} \\
MaskDiT                  & 1600 & 2.28                & 5.67          & 276.6          & 0.80          & 0.61       \\
DiT                      & 1400 & 2.27                & 4.60          & 278.2          & \textbf{0.83} & 0.51       \\
SiT                      & 1400 & 2.06                & 4.50 & 270.3          & 0.82          & 0.59       \\
FasterDiT                & 400  & 2.03                & 4.63          & 264.0          & 0.81          & 0.60       \\
MDT                      & 1300 & 1.79                & 4.57          & 283.0          & 0.81          & 0.61       \\
DiT+TREAD                & 740  & 1.69                & 4.73          & 292.7          & 0.81          & 0.63       \\
MDTv2                    & 1080 & 1.58                & 4.52          & \textbf{314.7} & 0.79          & \textbf{0.65} \\
SiT+REPA                 & 800  & \textbf{1.42} & 4.70          & 305.7          & 0.80          & \textbf{0.65} \\ 
\rowcolor{gray!8} 
\cellcolor{gray!8} & 100  & 1.74                & 4.74          & 268.7          & 0.80          & 0.62       \\
\rowcolor{gray!8} 
\cellcolor{gray!8} & 400  & 1.44                & 4.55          & 293.4          & 0.80          & 0.64       \\
\rowcolor{gray!8} 
\multirow{-3}{*}{\cellcolor{gray!8}SiT+\textbf{HASTE}} & 500    & \textbf{1.42}      & \textbf{4.49}             & 299.5        & 0.80            & \textbf{0.65}

\end{tabular}

\caption{System-level comparison on ImageNet 256 $\times$ 256. $\uparrow$ and $\downarrow$ denote higher and lower values are better, respectively. \textbf{Bold font} denotes the best performance.}
\label{tab:ext-draes-repa-xl}
\end{wraptable}

\paragraph{Results without classifier-free guidance.}
As shown in Table~\ref{tab:ext-draes-repa-xl}, HASTE demonstrates significant acceleration performance, consistently outperforming REPA on both SiT-XL and DiT-XL. This validates the superiority of stage-wise termination and holistic alignment. Notably, on SiT-XL, HASTE achieves an FID of 8.39 with only 250K iterations (50 epochs), matching the performance of vanilla SiT-XL with 1400 epochs, representing a 28$\times$ acceleration. Similarly, on DiT-XL, our approach surpasses the original DiT-XL trained with 1400 epochs, using only 80 epochs.

\paragraph{Results with classifier-free guidance.}
We also evaluate the generation performance of SiT-XL+HASTE at different epochs with classifier-free guidance (CFG) \citep{ho2022classifierfreediffusionguidance} applying guidance interval \citep{Kynkaanniemi2024}. As shown in Table~\ref{tab:ext-draes-repa-xl}, HASTE outperforms most of the baselines in only 400 epochs, and can achieve comparable FID score to REPA with 500 epochs, which proves that in later training stages, the denoising objective itself is also able to lead diffusion transformers to satisfactory generation capability.

\begin{figure}[htp]

  \centering
    \includegraphics[width=1.0\linewidth]{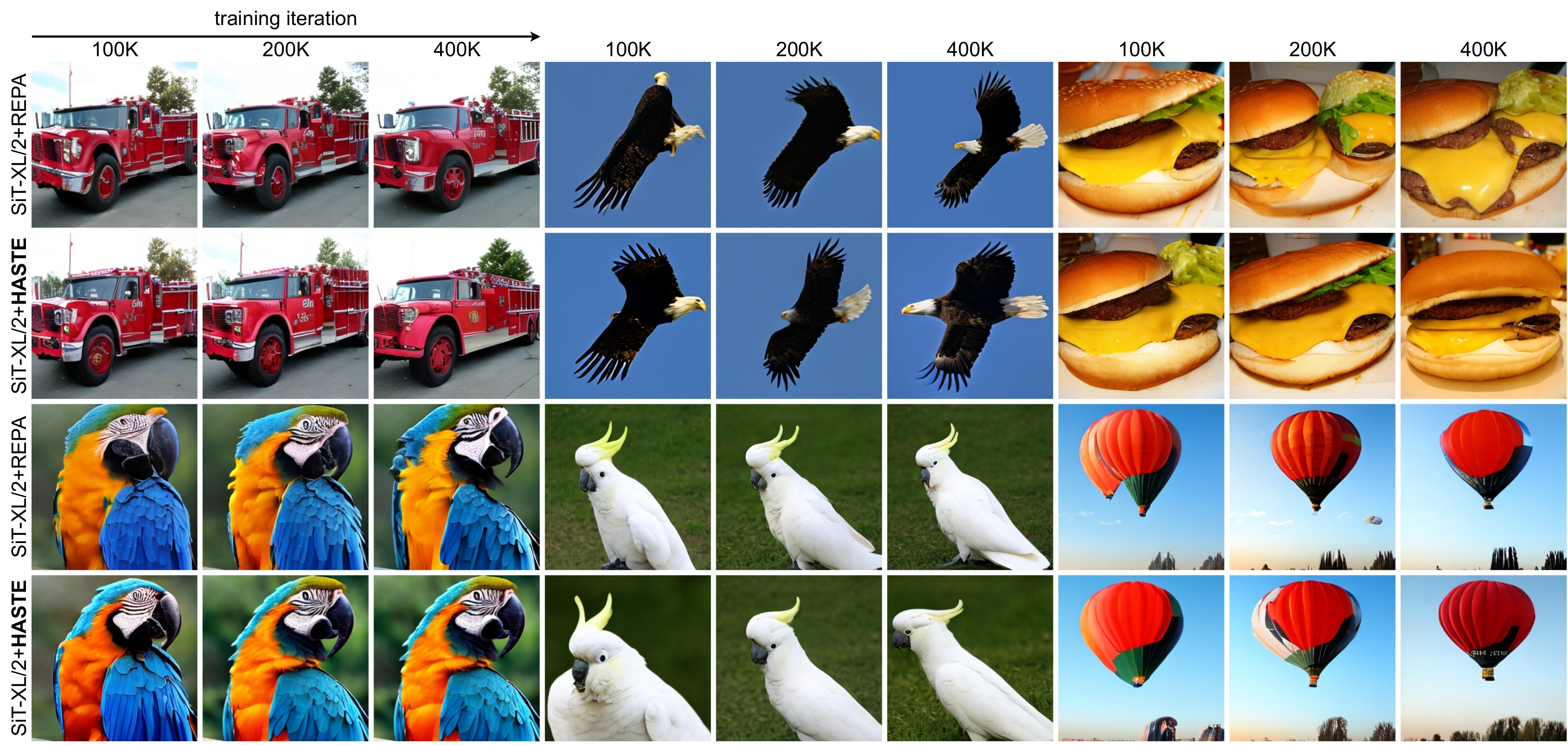}
   \caption{HASTE improves visual scaling. 
   We compare images generated by SiT-XL/2+REPA and SiT-XL/2+HASTE (ours) at different training iterations. 
   For both models, we use the same seed, noise, and sampling method with a classifier-free guidance scale of 4.0.}
    \label{fig:main-comparison}
\end{figure}

\paragraph{Qualitatively comparison.}
We also provide representative visualization results from SiT-XL/2 with REPA and HASTE in Figure~\ref{fig:main-comparison}, respectively. Our method achieves better semantic information and detail generation at early training stages.

\subsection{Text-to-Image Generation Experiment}

\paragraph{Setting.}
To validate our approach in text-to-image generation tasks, we apply HASTE to MM-DiT \citep{esser2024scalingrectifiedflowtransformers}, a widely used architecture, and train it on the MS-COCO 2014 dataset \citep{MSCOCO} for 150K iterations following REPA.
In practice, we do not apply alignment termination because of limited iteration number.
Moreover, we only perform attention alignment with the $QK^T$ matrix generated from input image to avoid affecting the textual process.

\paragraph{Quantitative results.}
In Table~\ref{tab:exp-t2i}, we compare our method with the original MM-DiT and MM-DiT+REPA using ODE and SDE samplers.Results reflect that our method consistently outperforms its counterparts, validating the generalizability of our holistic alignment in text-to-image generation.

\subsection{Ablation Studies} \label{AblationStudies}

\vspace{-1em}
In this section, we conduct extensive 
experiments and comparisons across different SiT models on ImageNet 256$\times$256, to further support our analysis and claims in Section~\ref{sec:Method}. We consistently use the SDE Euler-Maruyama sampler ($\text{NFEs} = 250$) without classifier-free guidance. 

\begin{wraptable}[12]{R}{0.50\textwidth}
\tablestyle{3pt}{1.3}
\centering
\begin{tabular}{lcccc}
\multirow{2}{*}{model} & \multicolumn{2}{c}{ODE (NFEs = 50)} & \multicolumn{2}{c}{SDE (NFEs = 250)} \\
\cmidrule(lr){2-3} \cmidrule(lr){4-5}       & w/o CFG   & w/ CFG & w/o CFG & w/ CFG         \\ \midrule[1.2pt]
MM-DiT  & 15.42     & 6.35   & 11.76   & 5.26          \\
+REPA  & 10.40 &   4.95     & 7.33    & 4.16          \\ \hline
\rowcolor{gray!8}
+\textbf{HASTE} & \textbf{9.63}     & \textbf{4.55}  & \textbf{6.81}   & \textbf{4.09}
\end{tabular}
  \caption{FID$\downarrow$ results of text-to-image generation on MS-COCO. Our holistic alignment method outperforms REPA in the early training stage.}
  \label{tab:exp-t2i}
\end{wraptable}

\paragraph{Effectiveness of ATTA and termination.} 
To validate the effectiveness of termination and Attention Alignment, we evaluate the performance of SiT-XL/2 with different methods applied before and after the termination point (50 epoch) and present the results in Table~\ref{tab:ablation-all}. 
Firstly, at both 40 and 100 epochs, we observe that using only Attention Alignment can also obtain a similar acceleration to REPA. 
Moreover, the holistic alignment leads to better performance at 40 epoch, which is consistent with our hypothesis in Section~\ref{sec:attn} that the two methods have complementary potentials.

However, the acceleration of such integration gets inferior to REPA alone at 100 epoch. We assume that consistently applying holistic alignment leads to over-regularization in later training stages. And  the performance gets improved eventually with the termination strategy applied at 50 epoch.

\paragraph{Different termination iterations $\tau$.}
In this section, we analyze the impact of $\tau$ across varying model sizes. First, we conduct experiments in Table~\ref{tab:es-b-l} to further explore the effect of termination. The results reflect that stage-wise termination also leads to better generation quality on SiT-B/2 and SiT-L/2. For SiT-XL/2, interestingly, while $\tau=400$ K demonstrates a lower FID at 400K iteration, $\tau=250$ K model ultimately delivers superior performance when evaluated at 500K iteration.

As shown in Table~\ref{tab:ablation-all} and Table~\ref{tab:es-b-l}, although holistic alignment achieves better performance at 400K iteration, consistently regularizing the model leads to reduced performance. 
While termination at $\tau=400$ K alleviates such a trend, its performance at 500K iteration is still inferior to that of $\tau=250$ K. Therefore, we hypothesize that the acceleration effect gradually diminishes before 400K iteration, and the stage-wise termination, such as at $\tau=250$ K, can help to alleviate the over-regularization.

\begin{minipage}[B]{0.49\textwidth}
\vspace{5pt}
\tablestyle{3pt}{1.3}
\centering
\begin{tabular}{ccccccc}
epoch                   & REPA     & ATTA     & ter.       & FID$\downarrow$ & sFID$\downarrow$ & IS$\uparrow$ \\ \midrule[1.2pt]
                         & $\times$ & $\times$ & $\times$ & 24.3            & 5.08             & 56.1         \\
\multirow{-2}{*}{40}     & $\circ$  & $\times$ & $\times$ & 10.7            & \textbf{5.02}    & 103.9        \\
\rowcolor{gray!8} 
\cellcolor{gray!8} & $\times$ & $\circ$  & $\times$ & 13.6            & \textbf{{5.02}}    & 89.7         \\
\rowcolor{gray!8} 
\multirow{-2}{*}{\cellcolor{gray!8}40}  & $\circ$ & $\circ$ & $\times$ & \textbf{9.9} & 5.04          & \textbf{108.8} \\ \hline
                         & $\times$ & $\times$ & $\times$ & 14.8            & 5.18             & 84.9         \\
\multirow{-2}{*}{100}    & $\circ$  & $\times$ & $\times$ & 7.5             & 5.11             & 130.1        \\
\rowcolor{gray!8} 
\cellcolor{gray!8} & $\times$ & $\circ$  & $\times$ & 8.5             & 5.00             & 120.7        \\
\rowcolor{gray!8} 
\cellcolor{gray!8} & $\circ$  & $\circ$  & $\times$ & 8.1             & 5.20             & 126.1        \\
\rowcolor{gray!8} 
\multirow{-3}{*}{\cellcolor{gray!8}100} & $\circ$ & $\circ$ & $\circ$  & \textbf{5.3} & \textbf{4.72} & \textbf{148.5}
\end{tabular}
  \captionof{table}{Comparison of different methods applied to SiT-XL/2. $\circ$ and $\times$ denote methods applied or not, respectively. Results reflect that our termination and holistic alignment strategies are effective.}
  \label{tab:ablation-all}
  \vspace{5pt}
\end{minipage}
\hfill
\begin{minipage}[B]{0.49\textwidth}
\vspace{5pt}
\tablestyle{3pt}{1.3}
\centering
\begin{tabular}{lccccc}
model                                                                     & iteration              & $\tau$           & FID$\downarrow$ & sFID$\downarrow$ & IS$\uparrow$   \\ \midrule[1.2pt]
\multirow{2}{*}{\begin{tabular}[c]{@{}l@{}}SiT-B/2\\  +\textbf{HASTE}\end{tabular}} & \multirow{2}{*}{400K} & - & 21.3            & 6.80             & 69.9           \\
                                                                          &                       & 100K             & \textbf{19.6}   & \textbf{6.38}    & \textbf{73.0}  \\ \hline
\multirow{2}{*}{\begin{tabular}[c]{@{}l@{}}SiT-L/2\\ +\textbf{HASTE}\end{tabular}} & \multirow{2}{*}{400K} & - & 8.9             & 5.18             & 119.0          \\
                                                                          &                       & 250K             & \textbf{7.9}    & \textbf{5.08}    & \textbf{124.8} \\ \hline
\multirow{2}{*}{\begin{tabular}[c]{@{}l@{}}SiT-XL/2\\ +\textbf{HASTE}\end{tabular}} & \multirow{2}{*}{400K} & - & \textbf{5.5} & \textbf{4.74} & \textbf{144.4} \\ 
& & 250K & 7.3 & 5.05 & 128.7 \\ \hline
\multirow{3}{*}{\begin{tabular}[c]{@{}l@{}}SiT-XL/2\\ +\textbf{HASTE}\end{tabular}}      & \multirow{3}{*}{500K} & - & 8.1             & 5.20             & 126.1          \\
      &                       & 250K             & \textbf{5.3}    & \textbf{4.72}    & \textbf{148.5} \\
      &                       & 400K             & 7.4             & 5.10             & 128.8         
\end{tabular}
    \captionof{table}{Comparison of applying termination or not across different model sizes of SiT. $\tau$ denotes termination point. We find the termination strategy contributes to better performance eventually. 
    }
  \label{tab:es-b-l}
\vspace{5pt}
\end{minipage}

\begin{wrapfigure}[15]{r}{0.5\textwidth}
\vspace{-2em}
  \centering
      \begin{subfigure}{0.45\linewidth}
        \includegraphics[width=1.0\linewidth]{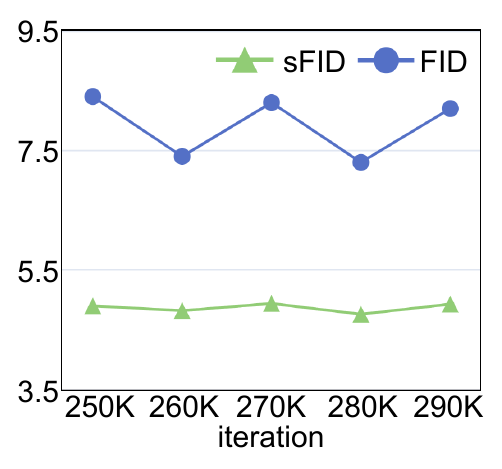}
        \caption{FID and sFID around 270K iter. on SiT-XL/2 without termination.}
        \label{fig:ES-iters}
      \end{subfigure}
      \hfill
      \begin{subfigure}{0.45\linewidth}
        \includegraphics[width=1.0\linewidth]{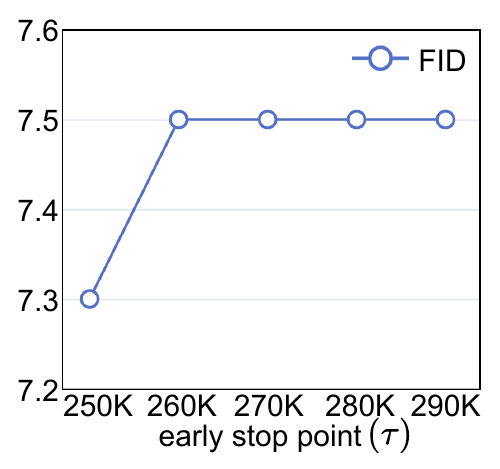}
        \caption{FID at 400K iter. with termination point $\tau$ around 270K.}
        \label{fig:ES-400K}
      \end{subfigure}
   \caption{Comparison of different termination point $\tau$ on SiT-XL/2. We observe the training oscillation after 250K iteration. Using $\tau=250$ K leads to better performance at 400K iteration.}
    \label{fig:ES-XL/2}  
\vspace{1em}
\end{wrapfigure}

Taking SiT-XL/2 for example, we carefully assess the effect of different $\tau$. We observe performance progresses slowly after 250K iteration (see Figure~\ref{fig:ES-iters}). And the gradient cosine similarity between holistic alignment and denoising has shown negative values at late diffusion timesteps (see details in Appendix~\ref{app:grad_ang}). Consequently, we consider termination near this threshold: results at 400K iteration in Figure~\ref{fig:ES-400K} indicate that early stopping at $\tau=250K$ yields better performance.

\paragraph{Different Attention Alignment loss weight $\lambda_{A}$.}
We evaluate the sensitivity of model to the attention alignment loss weight $\lambda_A$ in Equation~\ref{eq:hybrid} with SiT-L/2 as an example.

As shown in Table~\ref{tab:lambda_A_L}, HASTE consistently improves the performance of SiT-L/2 at 400K iteration across different values of $\lambda_{A}$, indicating that attention alignment provides relatively stable benefits. 
We note that larger weights can lead to reduced performance.
Therefore, we choose $\lambda_A=0.5$ in our primary experiments.

\noindent
\textbf{Selection of alignment layers}. 
We try different transfer layers for HASTE on SiT-L/2 in Table~\ref{tab:layer_choosing}. 
For brevity, we denote layers from SiT and DINO as layer-S and layer-D, respectively. 
Additionally, we use $[\cdot]_S$ and $[\cdot]_D$ to specify particular layer indices (counting from 0).

Firstly, we find that enough deeper layers should get involved for optimal performance. 
As shown in Table~\ref{tab:layer_choosing}, when choosing only two layers of each model for alignment, namely $[10, 11]_D$ and $[6, 7]_S$, the performance is inferior to choosing four layers. 
Additionally, results reflect that enough shallow layers should be left for processing the noisy inputs in the latent space. 
We can observe that the distillation of $[8,9,10,11]_D$ to $[4,5,6,7]_S$ achieves a better FID without including $[6, 7]_D$ to $[2,3]_S$.

\begin{minipage}[h]{0.34\textwidth} 
\tablestyle{2.8pt}{1.3}
\centering
\vspace{0.17em}
    \begin{tabular}{lccccc}
model &  $\lambda_A$ & FID$\downarrow$ & sFID$\downarrow$ & IS$\uparrow$ \\ \midrule[1.2pt]
\multirow{4}{*}{\begin{tabular}[c]{@{}c@{}}SiT-L/2\\+\textbf{HASTE}\end{tabular}} &  0.5 & \textbf{7.9} & \textbf{5.08} & \textbf{124.8} \\
             & 1.0         & 8.6             & 5.29             & 120.0        \\
             & 1.5         & 8.7             & 5.23             & 119.3        \\
             & 3.0         & 9.0             & 5.34             & 116.9       
\end{tabular}
\vspace{-0.17em}
  \captionof{table}{ATTA weight $\lambda_A=0.5$ leads to better performance on SiT-L/2 at 400K iteration.
  }
  \label{tab:lambda_A_L}
\end{minipage}
\hfill
 \begin{minipage}[h]{0.64\textwidth}
     \centering
\tablestyle{2.5pt}{1.3}
\begin{tabular}{lccccc}
model & layer-D             & layer-S           & FID$\downarrow$ & sFID$\downarrow$ & IS$\uparrow$   \\ \midrule[1.2pt]
\multirow{3}{*}{\begin{tabular}[c]{@{}c@{}}SiT-L/2\\ +\textbf{HASTE}\end{tabular}} & {$[10,11]_D$} & {$[6,7]_S$} & 8.9 & 5.31 & 119.3 \\
      & {$[8,9,10,11]_D$}     & {$[4,5,6,7]_S$}     & \textbf{7.9}    & \textbf{5.08}    & \textbf{124.8} \\
      & {$[6,7,8,9,10,11]_D$} & {$[2,3,4,5,6,7]_S$} & 8.3             & 5.12             & 121.3         
\end{tabular}
  \captionof{table}{Comparison of HASTE with different choices of layers on SiT-L/2 at 400K iteration. While transferring attention maps for more deep layers provides greater benefits, we need to preserve enough shallow layers to process latent input.}
  \label{tab:layer_choosing}
 \end{minipage}

Our findings align with the observations of attention transfer on ViTs reported in \citep{li2024on}: transferring more attention maps from deeper layers provides greater benefits, and ViTs can learn low-level features well when guided on how to integrate these features into higher-level ones.

\section{Related Work}
\subsection{Accelerating Training Diffusion Transformers}
To accelerate the training of diffusion transformers, existing methods can be broadly classified into two categories: architectural modifications and representation enhancements.

\paragraph{Architecture modification.}
These methods focus on directly improving the efficiency of the model architecture. 
For example, SANA series \citep{xie2024sana, xie2025sana15efficientscaling}, DiG \citep{DBLP:journals/corr/abs-2405-18428}, and LiT \citep{wang2025litdelvingsimplifiedlinear} introduce Linear Attention \citep{10.5555/3524938.3525416, wang2020linformer, choromanski2021rethinking} to improve the efficiency of diffusion transformers. 
Additionally, methods like MaskDiT \citep{zheng2024fast} and MDT \citep{Gao2023MaskedDT, gao2024mdtv2maskeddiffusiontransformer} introduce masked image modeling~\cite{he2022masked} to reduce the cost during training.

\paragraph{Representation incorporation.} 
In contrast to architecture modifications, these methods do not require designing specialized structures and instead leverage external representations to achieve acceleration.  
For instance, REPA~\citep{yu2025repa} observes the difficulty in learning effective representations for diffusion models~\citep{sohl2015deep, ho2020denoising, song2021scorebased}, which hinders the training efficiency. 
To address this, REPA proposes to align the internal features of diffusion transformers with the output of pre-trained representation models, and significantly accelerates the training process. 

Furthermore, recent works~\citep{vavae, tian2025urepaaligningdiffusionunets, leng2025repaeunlockingvaeendtoend} have also achieved better results based on representation methods. 
For example, U-REPA~\citep{tian2025urepaaligningdiffusionunets} improves REPA with a manifold alignment loss, and demonstrates its effectiveness on U-Nets~\citep{U-Nets}.
External representations can also help enhance generation and reconstruction capabilities of VAE, such as in VA-VAE~\citep{vavae} and E2E-VAE~\citep{leng2025repaeunlockingvaeendtoend}.

Unlike these methods, our research focuses mainly on the diffusion transformer itself. We investigate the relationship between external representation guidance and the self-improvement of diffusion transformers, and propose to remove the regularization at an appropriate training stage.

\subsection{Attention Transfer for Vision Transformers}

The attention mechanism \citep{attention} has been shown to provide vision models, such as Vision Transformers (ViTs) \citep{dosovitskiy2021an}, with strong adaptability and scalability across various tasks. While prior works \citep{MAE, Moco} have achieved improved downstream performance by leveraging entire pre-trained models, \citet{li2024on} demonstrates that the attention patterns learned during pre-training are sufficient for ViTs to learn high-quality representations from scratch, achieving performance comparable to fine-tuned models on downstream tasks. Consequently, attention distillation \citep{li2024on, Wang0022-1, ren2023tinymim} has been proposed to transfer knowledge efficiently.

The transfer of attention maps has been extensively studied in Vision Transformers (ViTs), but remains underexplored in diffusion transformers. 
While recent work \citep{zhou2025attentiondistillationunifiedapproach} applies attention distillation for characteristics transfer tasks using diffusion models, its explorations remain in the sampling process. 
Moreover, the relationship between attention mechanisms in ViTs and diffusion transformers requires further investigation.  In this work, we demonstrate that attention maps from a pre-trained ViT can effectively guide the learning process of diffusion transformers.

\section{Conclusion}

 In this paper, we have proposed HASTE, a simple but effective way to improve the training efficiency of diffusion transformers. Specifically, we reveal that \emph{representation alignment is not always beneficial throughout the training process}.
 In addition, we analyze the stages when feature alignment is most effective and investigate the dilemma between external feature guidance and internal self-improvement of diffusion transformers. We prove that HASTE can significantly accelerate the training process of mainstream diffusion transformers, such as SiT and DiT. 
 We hope our work would further reduce the cost for researchers to train diffusion transformers, and the application of diffusion models in downstream tasks.

\paragraph{Limitations and future work.} We mainly focus on diffusion transformers in latent space for image generation. 
Explorations of HASTE with pixel-level diffusion \citep{ADM,VDM}, or in video generation tasks \citep{ho2022video} would be exciting directions for future work. 
Additionally, HASTE may also be incorporated with other methods \citep{vavae, leng2025repaeunlockingvaeendtoend} on different model architectures \citep{tian2025urepaaligningdiffusionunets}. 

\paragraph{Ackonwledgement.} We sincerely appreciate Liang Zheng and Ziheng Qin for valuable discussions and feedbacks during this work.

{
    \small
    \bibliographystyle{plainnat}
    \bibliography{neurips_2025}

\begin{thebibliography}{58}
\providecommand{\natexlab}[1]{#1}
\providecommand{\url}[1]{\texttt{#1}}
\expandafter\ifx\csname urlstyle\endcsname\relax
  \providecommand{\doi}[1]{doi: #1}\else
  \providecommand{\doi}{doi: \begingroup \urlstyle{rm}\Url}\fi

\bibitem[Bao et~al.(2023)Bao, Nie, Xue, Cao, Li, Su, and Zhu]{bao2022all}
Fan Bao, Shen Nie, Kaiwen Xue, Yue Cao, Chongxuan Li, Hang Su, and Jun Zhu.
\newblock All are worth words: A vit backbone for diffusion models.
\newblock In \emph{CVPR}, 2023.

\bibitem[Brooks et~al.(2024)Brooks, Peebles, Holmes, DePue, Guo, Jing, Schnurr, Taylor, Luhman, Luhman, Ng, Wang, and Ramesh]{videoworldsimulators2024}
Tim Brooks, Bill Peebles, Connor Holmes, Will DePue, Yufei Guo, Li~Jing, David Schnurr, Joe Taylor, Troy Luhman, Eric Luhman, Clarence Ng, Ricky Wang, and Aditya Ramesh.
\newblock Video generation models as world simulators.
\newblock https://openai.com/research/video-generation-models-as-world-simulators, 2024.

\bibitem[Choromanski et~al.(2021)Choromanski, Likhosherstov, Dohan, Song, Gane, Sarlos, Hawkins, Davis, Mohiuddin, Kaiser, Belanger, Colwell, and Weller]{choromanski2021rethinking}
Krzysztof~Marcin Choromanski, Valerii Likhosherstov, David Dohan, Xingyou Song, Andreea Gane, Tamas Sarlos, Peter Hawkins, Jared~Quincy Davis, Afroz Mohiuddin, Lukasz Kaiser, David~Benjamin Belanger, Lucy~J Colwell, and Adrian Weller.
\newblock Rethinking attention with performers.
\newblock In \emph{ICLR}, 2021.

\bibitem[Deng et~al.(2009)Deng, Dong, Socher, Li, Li, and Fei-Fei]{ImageNet}
Jia Deng, Wei Dong, Richard Socher, Li-Jia Li, Kai Li, and Li~Fei-Fei.
\newblock Imagenet: A large-scale hierarchical image database.
\newblock In \emph{CVPR}, 2009.

\bibitem[Dhariwal and Nichol(2021)]{ADM}
Prafulla Dhariwal and Alex Nichol.
\newblock Diffusion models beat gans on image synthesis.
\newblock In \emph{NeurIPS}, 2021.

\bibitem[Dosovitskiy et~al.(2021)Dosovitskiy, Beyer, Kolesnikov, Weissenborn, Zhai, Unterthiner, Dehghani, Minderer, Heigold, Gelly, Uszkoreit, and Houlsby]{dosovitskiy2021an}
Alexey Dosovitskiy, Lucas Beyer, Alexander Kolesnikov, Dirk Weissenborn, Xiaohua Zhai, Thomas Unterthiner, Mostafa Dehghani, Matthias Minderer, Georg Heigold, Sylvain Gelly, Jakob Uszkoreit, and Neil Houlsby.
\newblock An image is worth 16x16 words: Transformers for image recognition at scale.
\newblock In \emph{ICLR}, 2021.

\bibitem[Elfwing et~al.(2018)Elfwing, Uchibe, and Doya]{SiLu2018}
Stefan Elfwing, Eiji Uchibe, and Kenji Doya.
\newblock Sigmoid-weighted linear units for neural network function approximation in reinforcement learning.
\newblock \emph{Neural Networks}, 107:\penalty0 3--11, 2018.
\newblock ISSN 0893-6080.
\newblock \doi{https://doi.org/10.1016/j.neunet.2017.12.012}.
\newblock URL \url{https://www.sciencedirect.com/science/article/pii/S0893608017302976}.
\newblock Special issue on deep reinforcement learning.

\bibitem[Esser et~al.(2024)Esser, Kulal, Blattmann, Entezari, M{\"u}ller, Saini, Levi, Lorenz, Sauer, Boesel, et~al.]{esser2024scalingrectifiedflowtransformers}
Patrick Esser, Sumith Kulal, Andreas Blattmann, Rahim Entezari, Jonas M{\"u}ller, Harry Saini, Yam Levi, Dominik Lorenz, Axel Sauer, Frederic Boesel, et~al.
\newblock Scaling rectified flow transformers for high-resolution image synthesis.
\newblock In \emph{ICML}, 2024.

\bibitem[Feng et~al.(2025)Feng, Ma, Wang, Qi, Chen, Chen, and Wang]{feng2024dit4edit}
Kunyu Feng, Yue Ma, Bingyuan Wang, Chenyang Qi, Haozhe Chen, Qifeng Chen, and Zeyu Wang.
\newblock Dit4edit: Diffusion transformer for image editing.
\newblock In \emph{AAAI}, 2025.

\bibitem[Gao et~al.(2023)Gao, Zhou, Cheng, and Yan]{Gao2023MaskedDT}
Shanghua Gao, Pan Zhou, Mingg-Ming Cheng, and Shuicheng Yan.
\newblock Masked diffusion transformer is a strong image synthesizer.
\newblock In \emph{ICCV}, 2023.

\bibitem[Gao et~al.(2024)Gao, Zhou, Cheng, and Yan]{gao2024mdtv2maskeddiffusiontransformer}
Shanghua Gao, Pan Zhou, Ming-Ming Cheng, and Shuicheng Yan.
\newblock Mdtv2: Masked diffusion transformer is a strong image synthesizer.
\newblock \emph{arXiv preprint arXiv:2303.14389}, 2024.
\newblock URL \url{https://arxiv.org/abs/2303.14389}.

\bibitem[Hang et~al.(2023)Hang, Gu, Li, Bao, Chen, Hu, Geng, and Guo]{Hang_2023_ICCV}
Tiankai Hang, Shuyang Gu, Chen Li, Jianmin Bao, Dong Chen, Han Hu, Xin Geng, and Baining Guo.
\newblock Efficient diffusion training via min-snr weighting strategy.
\newblock In \emph{ICCV}, 2023.

\bibitem[He et~al.(2020)He, Fan, Wu, Xie, and Girshick]{Moco}
Kaiming He, Haoqi Fan, Yuxin Wu, Saining Xie, and Ross Girshick.
\newblock Momentum contrast for unsupervised visual representation learning.
\newblock In \emph{CVPR}, 2020.

\bibitem[He et~al.(2022{\natexlab{a}})He, Chen, Xie, Li, Doll{\'a}r, and Girshick]{he2022masked}
Kaiming He, Xinlei Chen, Saining Xie, Yanghao Li, Piotr Doll{\'a}r, and Ross Girshick.
\newblock Masked autoencoders are scalable vision learners.
\newblock In \emph{CVPR}, 2022{\natexlab{a}}.

\bibitem[He et~al.(2022{\natexlab{b}})He, Chen, Xie, Li, Dollár, and Girshick]{MAE}
Kaiming He, Xinlei Chen, Saining Xie, Yanghao Li, Piotr Dollár, and Ross Girshick.
\newblock Masked autoencoders are scalable vision learners.
\newblock In \emph{CVPR}, 2022{\natexlab{b}}.

\bibitem[Heusel et~al.(2017)Heusel, Ramsauer, Unterthiner, Nessler, and Hochreiter]{10.5555/3295222.3295408}
Martin Heusel, Hubert Ramsauer, Thomas Unterthiner, Bernhard Nessler, and Sepp Hochreiter.
\newblock Gans trained by a two time-scale update rule converge to a local nash equilibrium.
\newblock In \emph{NeurIPS}, 2017.

\bibitem[Ho and Salimans(2021)]{ho2022classifierfreediffusionguidance}
Jonathan Ho and Tim Salimans.
\newblock Classifier-free diffusion guidance.
\newblock In \emph{NeurIPS Workshop}, 2021.

\bibitem[Ho et~al.(2020)Ho, Jain, and Abbeel]{ho2020denoising}
Jonathan Ho, Ajay Jain, and Pieter Abbeel.
\newblock Denoising diffusion probabilistic models.
\newblock In \emph{NeurIPS}, 2020.

\bibitem[Ho et~al.(2022)Ho, Salimans, Gritsenko, Chan, Norouzi, and Fleet]{ho2022video}
Jonathan Ho, Tim Salimans, Alexey Gritsenko, William Chan, Mohammad Norouzi, and David~J Fleet.
\newblock Video diffusion models.
\newblock In \emph{NeurIPS}, 2022.

\bibitem[Hoang et~al.(2024)Hoang, Cho, Merth, Wang, Rastegari, and Naik]{hoang2024compressedllmsforgetknowledge}
Scott Hoang, Minsik Cho, Thomas Merth, Atlas Wang, Mohammad Rastegari, and Devang Naik.
\newblock Do compressed llms forget knowledge? an experimental study with practical implications.
\newblock In \emph{NeurIPS Workshop}, 2024.

\bibitem[Karras et~al.(2024)Karras, Aittala, Lehtinen, Hellsten, Aila, and Laine]{Karras2024edm2}
Tero Karras, Miika Aittala, Jaakko Lehtinen, Janne Hellsten, Timo Aila, and Samuli Laine.
\newblock Analyzing and improving the training dynamics of diffusion models.
\newblock In \emph{CVPR}, 2024.

\bibitem[Katharopoulos et~al.(2020)Katharopoulos, Vyas, Pappas, and Fleuret]{10.5555/3524938.3525416}
Angelos Katharopoulos, Apoorv Vyas, Nikolaos Pappas, and Fran\c{c}ois Fleuret.
\newblock Transformers are rnns: fast autoregressive transformers with linear attention.
\newblock In \emph{ICML}, 2020.

\bibitem[Kim et~al.(2022)Kim, Shin, Song, Kang, and Moon]{DBLP:conf/icml/KimSSKM22}
Dongjun Kim, Seungjae Shin, Kyungwoo Song, Wanmo Kang, and Il-Chul Moon.
\newblock Soft truncation: A universal training technique of score-based diffusion model for high precision score estimation.
\newblock In \emph{ICML}, 2022.

\bibitem[Kingma and Gao(2023)]{VDM}
Diederik Kingma and Ruiqi Gao.
\newblock Understanding diffusion objectives as the elbo with simple data augmentation.
\newblock In \emph{NeurIPS}, 2023.

\bibitem[Kingma and Ba(2015)]{kingma2017adammethodstochasticoptimization}
Diederik~P. Kingma and Jimmy Ba.
\newblock Adam: A method for stochastic optimization.
\newblock In \emph{ICLR}, 2015.

\bibitem[Krause et~al.(2025)Krause, Phan, Hu, and Ommer]{krause2025treadtokenroutingefficient}
Felix Krause, Timy Phan, Vincent~Tao Hu, and Björn Ommer.
\newblock Tread: Token routing for efficient architecture-agnostic diffusion training.
\newblock \emph{arXiv preprint arXiv:2501.04765}, 2025.

\bibitem[Kynk\"{a}\"{a}nniemi et~al.(2019)Kynk\"{a}\"{a}nniemi, Karras, Laine, Lehtinen, and Aila]{precandrec}
Tuomas Kynk\"{a}\"{a}nniemi, Tero Karras, Samuli Laine, Jaakko Lehtinen, and Timo Aila.
\newblock Improved precision and recall metric for assessing generative models.
\newblock In \emph{NeurIPS}, 2019.

\bibitem[Kynkäänniemi et~al.(2024)Kynkäänniemi, Aittala, Karras, Laine, Aila, and Lehtinen]{Kynkaanniemi2024}
Tuomas Kynkäänniemi, Miika Aittala, Tero Karras, Samuli Laine, Timo Aila, and Jaakko Lehtinen.
\newblock Applying guidance in a limited interval improves sample and distribution quality in diffusion models.
\newblock In \emph{NeurIPS}, 2024.

\bibitem[Labs(2024)]{flux2024}
Black~Forest Labs.
\newblock Flux.
\newblock \url{https://github.com/black-forest-labs/flux}, 2024.

\bibitem[Leng et~al.(2025)Leng, Singh, Hou, Xing, Xie, and Zheng]{leng2025repaeunlockingvaeendtoend}
Xingjian Leng, Jaskirat Singh, Yunzhong Hou, Zhenchang Xing, Saining Xie, and Liang Zheng.
\newblock Repa-e: Unlocking vae for end-to-end tuning with latent diffusion transformers.
\newblock \emph{arXiv preprint arXiv:2504.10483}, 2025.
\newblock URL \url{https://arxiv.org/abs/2504.10483}.

\bibitem[Li et~al.(2024)Li, Tian, Chen, Pathak, and Chen]{li2024on}
Alexander~Cong Li, Yuandong Tian, Beidi Chen, Deepak Pathak, and Xinlei Chen.
\newblock On the surprising effectiveness of attention transfer for vision transformers.
\newblock In \emph{NeurIPS}, 2024.

\bibitem[Lin et~al.(2014)Lin, Maire, Belongie, Hays, Perona, Ramanan, Doll{\'a}r, and Zitnick]{MSCOCO}
Tsung-Yi Lin, Michael Maire, Serge Belongie, James Hays, Pietro Perona, Deva Ramanan, Piotr Doll{\'a}r, and C.~Lawrence Zitnick.
\newblock Microsoft coco: Common objects in context.
\newblock In \emph{ECCV}, 2014.

\bibitem[Loshchilov and Hutter(2017)]{loshchilov2019decoupledweightdecayregularization}
Ilya Loshchilov and Frank Hutter.
\newblock Decoupled weight decay regularization.
\newblock In \emph{ICLR}, 2017.

\bibitem[Ma et~al.(2024)Ma, Goldstein, Albergo, Boffi, Vanden-Eijnden, and Xie]{ma2024sitexploringflowdiffusionbased}
Nanye Ma, Mark Goldstein, Michael~S. Albergo, Nicholas~M. Boffi, Eric Vanden-Eijnden, and Saining Xie.
\newblock Sit: Exploring flow and diffusion-based generative models with scalable interpolant transformers.
\newblock \emph{arXiv preprint arXiv:2401.08740}, 2024.
\newblock URL \url{https://arxiv.org/abs/2401.08740}.

\bibitem[Nash et~al.(2021)Nash, Menick, Dieleman, and Battaglia]{pmlr-v139-nash21a}
Charlie Nash, Jacob Menick, Sander Dieleman, and Peter Battaglia.
\newblock Generating images with sparse representations.
\newblock In \emph{ICML}, 2021.

\bibitem[Oquab et~al.(2024)Oquab, Darcet, Moutakanni, Vo, Szafraniec, Khalidov, Fernandez, HAZIZA, Massa, El-Nouby, Assran, Ballas, Galuba, Howes, Huang, Li, Misra, Rabbat, Sharma, Synnaeve, Xu, Jegou, Mairal, Labatut, Joulin, and Bojanowski]{oquab2024dinov}
Maxime Oquab, Timoth{\'e}e Darcet, Th{\'e}o Moutakanni, Huy~V. Vo, Marc Szafraniec, Vasil Khalidov, Pierre Fernandez, Daniel HAZIZA, Francisco Massa, Alaaeldin El-Nouby, Mido Assran, Nicolas Ballas, Wojciech Galuba, Russell Howes, Po-Yao Huang, Shang-Wen Li, Ishan Misra, Michael Rabbat, Vasu Sharma, Gabriel Synnaeve, Hu~Xu, Herve Jegou, Julien Mairal, Patrick Labatut, Armand Joulin, and Piotr Bojanowski.
\newblock {DINO}v2: Learning robust visual features without supervision.
\newblock \emph{TMLR}, 2024.
\newblock ISSN 2835-8856.
\newblock URL \url{https://openreview.net/forum?id=a68SUt6zFt}.
\newblock Featured Certification.

\bibitem[Peebles and Xie(2023)]{Peebles2022DiT}
William Peebles and Saining Xie.
\newblock Scalable diffusion models with transformers.
\newblock In \emph{ICCV}, 2023.

\bibitem[Radford et~al.(2021)Radford, Kim, Hallacy, Ramesh, Goh, Agarwal, Sastry, Askell, Mishkin, Clark, Krueger, and Sutskever]{radford2021learning}
Alec Radford, Jong~Wook Kim, Chris Hallacy, Aditya Ramesh, Gabriel Goh, Sandhini Agarwal, Girish Sastry, Amanda Askell, Pamela Mishkin, Jack Clark, Gretchen Krueger, and Ilya Sutskever.
\newblock Learning transferable visual models from natural language supervision.
\newblock In \emph{ICML}, 2021.

\bibitem[Ren et~al.(2023)Ren, Wei, Zhang, and Hu]{ren2023tinymim}
Sucheng Ren, Fangyun Wei, Zheng Zhang, and Han Hu.
\newblock Tinymim: An empirical study of distilling mim pre-trained models.
\newblock In \emph{CVPR}, 2023.

\bibitem[Rombach et~al.(2022)Rombach, Blattmann, Lorenz, Esser, and Ommer]{LDM}
Robin Rombach, Andreas Blattmann, Dominik Lorenz, Patrick Esser, and Bj{\"{o}}rn Ommer.
\newblock High-resolution image synthesis with latent diffusion models.
\newblock In \emph{CVPR}, 2022.

\bibitem[Ronneberger et~al.(2015)Ronneberger, Fischer, and Brox]{U-Nets}
Olaf Ronneberger, Philipp Fischer, and Thomas Brox.
\newblock U-net: Convolutional networks for biomedical image segmentation.
\newblock In \emph{MICCAI}, 2015.

\bibitem[Salimans et~al.(2016)Salimans, Goodfellow, Zaremba, Cheung, Radford, and Chen]{IS}
Tim Salimans, Ian Goodfellow, Wojciech Zaremba, Vicki Cheung, Alec Radford, and Xi~Chen.
\newblock Improved techniques for training gans.
\newblock In \emph{NeurIPS}, 2016.

\bibitem[Sohl-Dickstein et~al.(2015)Sohl-Dickstein, Weiss, Maheswaranathan, and Ganguli]{sohl2015deep}
Jascha Sohl-Dickstein, Eric Weiss, Niru Maheswaranathan, and Surya Ganguli.
\newblock Deep unsupervised learning using nonequilibrium thermodynamics.
\newblock In \emph{ICML}, 2015.

\bibitem[Song et~al.(2021)Song, Sohl-Dickstein, Kingma, Kumar, Ermon, and Poole]{song2021scorebased}
Yang Song, Jascha Sohl-Dickstein, Diederik~P Kingma, Abhishek Kumar, Stefano Ermon, and Ben Poole.
\newblock Score-based generative modeling through stochastic differential equations.
\newblock In \emph{ICLR}, 2021.

\bibitem[Tian et~al.(2025)Tian, Chen, Zheng, Liang, Xu, and Wang]{tian2025urepaaligningdiffusionunets}
Yuchuan Tian, Hanting Chen, Mengyu Zheng, Yuchen Liang, Chao Xu, and Yunhe Wang.
\newblock U-repa: Aligning diffusion u-nets to vits.
\newblock \emph{arXiv preprint arXiv:2503.18414}, 2025.
\newblock URL \url{https://arxiv.org/abs/2503.18414}.

\bibitem[Vaswani et~al.(2017)Vaswani, Shazeer, Parmar, Uszkoreit, Jones, Gomez, Kaiser, and Polosukhin]{attention}
Ashish Vaswani, Noam Shazeer, Niki Parmar, Jakob Uszkoreit, Llion Jones, Aidan~N Gomez, \L~ukasz Kaiser, and Illia Polosukhin.
\newblock Attention is all you need.
\newblock In \emph{NeurIPS}, 2017.

\bibitem[Wang et~al.(2025)Wang, Kang, Yao, Chen, Wu, Zhang, Xue, Liu, Wu, Liu, Zhang, Zhang, Shao, Li, and Luo]{wang2025litdelvingsimplifiedlinear}
Jiahao Wang, Ning Kang, Lewei Yao, Mengzhao Chen, Chengyue Wu, Songyang Zhang, Shuchen Xue, Yong Liu, Taiqiang Wu, Xihui Liu, Kaipeng Zhang, Shifeng Zhang, Wenqi Shao, Zhenguo Li, and Ping Luo.
\newblock Lit: Delving into a simplified linear diffusion transformer for image generation.
\newblock \emph{arXiv preprint arXiv:2501.12976}, 2025.
\newblock URL \url{https://arxiv.org/abs/2501.12976}.

\bibitem[Wang et~al.(2022)Wang, 0004, and van~de Weijer~0001]{Wang0022-1}
Kai Wang, Fei~Yang 0004, and Joost van~de Weijer~0001.
\newblock Attention distillation: self-supervised vision transformer students need more guidance.
\newblock In \emph{BMVC}, 2022.

\bibitem[Wang et~al.(2024)Wang, Shi, Zhou, Li, Yuan, Shang, Peng, Zhang, and You]{wang2024closer}
Kai Wang, Mingjia Shi, Yukun Zhou, Zekai Li, Zhihang Yuan, Yuzhang Shang, Xiaojiang Peng, Hanwang Zhang, and Yang You.
\newblock A closer look at time steps is worthy of triple speed-up for diffusion model training.
\newblock \emph{arXiv preprint arXiv:2405.17403}, 2024.

\bibitem[Wang et~al.(2020)Wang, Li, Khabsa, Fang, and Ma]{wang2020linformer}
Sinong Wang, Belinda Li, Madian Khabsa, Han Fang, and Hao Ma.
\newblock Linformer: Self-attention with linear complexity.
\newblock \emph{arXiv preprint arXiv:2006.04768}, 2020.

\bibitem[Xie et~al.(2024)Xie, Chen, Chen, Cai, Tang, Lin, Zhang, Li, Zhu, Lu, and Han]{xie2024sana}
Enze Xie, Junsong Chen, Junyu Chen, Han Cai, Haotian Tang, Yujun Lin, Zhekai Zhang, Muyang Li, Ligeng Zhu, Yao Lu, and Song Han.
\newblock Sana: Efficient high-resolution image synthesis with linear diffusion transformer.
\newblock \emph{arXiv preprint arXiv:2410.10629}, 2024.
\newblock URL \url{https://arxiv.org/abs/2410.10629}.

\bibitem[Xie et~al.(2025)Xie, Chen, Zhao, Yu, Zhu, Lin, Zhang, Li, Chen, Cai, Liu, Zhou, and Han]{xie2025sana15efficientscaling}
Enze Xie, Junsong Chen, Yuyang Zhao, Jincheng Yu, Ligeng Zhu, Yujun Lin, Zhekai Zhang, Muyang Li, Junyu Chen, Han Cai, Bingchen Liu, Daquan Zhou, and Song Han.
\newblock Sana 1.5: Efficient scaling of training-time and inference-time compute in linear diffusion transformer.
\newblock \emph{arXiv preprint arXiv:2501.18427}, 2025.
\newblock URL \url{https://arxiv.org/abs/2501.18427}.

\bibitem[Yao and Wang(2025)]{vavae}
Jingfeng Yao and Xinggang Wang.
\newblock Reconstruction vs. generation: Taming optimization dilemma in latent diffusion models.
\newblock In \emph{CVPR}, 2025.

\bibitem[Yao et~al.(2024)Yao, Wang, Liu, and Wang]{fasterdit}
Jingfeng Yao, Cheng Wang, Wenyu Liu, and Xinggang Wang.
\newblock Fasterdit: Towards faster diffusion transformers training without architecture modification.
\newblock In \emph{NeurIPS}, 2024.

\bibitem[Yu et~al.(2025)Yu, Kwak, Jang, Jeong, Huang, Shin, and Xie]{yu2025repa}
Sihyun Yu, Sangkyung Kwak, Huiwon Jang, Jongheon Jeong, Jonathan Huang, Jinwoo Shin, and Saining Xie.
\newblock Representation alignment for generation: Training diffusion transformers is easier than you think.
\newblock In \emph{ICLR}, 2025.

\bibitem[Zheng et~al.(2024)Zheng, Nie, Vahdat, and Anandkumar]{zheng2024fast}
Hongkai Zheng, Weili Nie, Arash Vahdat, and Anima Anandkumar.
\newblock Fast training of diffusion models with masked transformers.
\newblock \emph{TMLR}, 2024.
\newblock ISSN 2835-8856.
\newblock URL \url{https://openreview.net/forum?id=vTBjBtGioE}.

\bibitem[Zhou et~al.(2025)Zhou, Gao, Chen, and Huang]{zhou2025attentiondistillationunifiedapproach}
Yang Zhou, Xu~Gao, Zichong Chen, and Hui Huang.
\newblock Attention distillation: A unified approach to visual characteristics transfer.
\newblock \emph{arXiv preprint arXiv:2502.20235}, 2025.
\newblock URL \url{https://arxiv.org/abs/2502.20235}.

\bibitem[Zhu et~al.(2024)Zhu, Huang, Liao, Liew, Yan, Feng, and Wang]{DBLP:journals/corr/abs-2405-18428}
Lianghui Zhu, Zilong Huang, Bencheng Liao, Jun~Hao Liew, Hanshu Yan, Jiashi Feng, and Xinggang Wang.
\newblock Dig: Scalable and efficient diffusion models with gated linear attention.
\newblock \emph{arXiv preprint arXiv:2405.18428}, 2024.
\newblock URL \url{https://arxiv.org/abs/2405.18428}.

\end{thebibliography}
}
\newpage
\appendix

\section{Additional Results}

\subsection{Gradient Angle}
\label{app:grad_ang}
We provide detailed results of cosine similarity between REPA~\citep{yu2025repa} and denoising gradients. 
In Figure~\ref{fig:app_grad_sim}, we separately compute gradients of the feature alignment and the denoising objective for SiT-XL/2~\citep{ma2024sitexploringflowdiffusionbased} and compare the cosine similarity of their directions at different training iterations.
Specifically, we randomly sample 960 images from the training dataset of ImageNet \citep{ImageNet} for the comparison and take gradients of parameters in the eighth block of SiT-XL/2 for example (REPA sets the default alignment depth as 8).

\begin{minipage}[h]{0.49\textwidth}
\vspace{1em}
  \centering
  \includegraphics[width=1\linewidth]{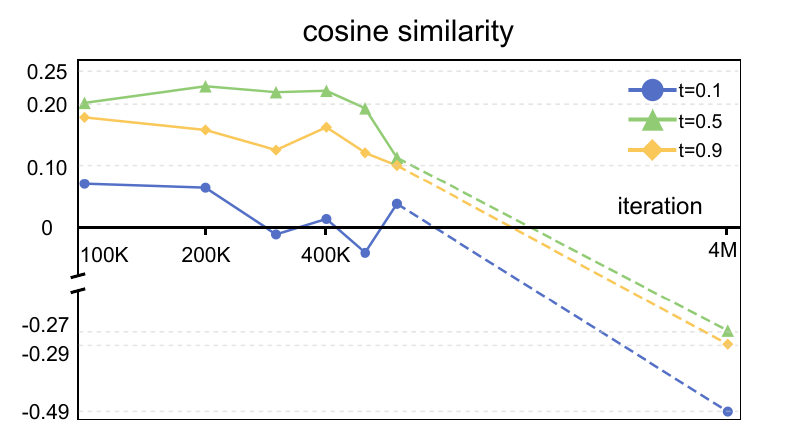}
  \captionof{figure}{Gradient cosine similarity between REPA and the denoising objective.}
  \label{fig:app_grad_sim}
  \vspace{1em}
\end{minipage}
\hfill
\begin{minipage}[h]{0.49\textwidth}
\vspace{1em}
\tablestyle{2.pt}{1.3}
  \centering
  \begin{tabular}{lccccc}
iteration & t = 0.02 & t = 0.04 & t = 0.06 & t = 0.08 & t = 0.10 \\ \midrule[1.2pt]
100K      & 0.0070   & 0.0064   & 0.0327   & 0.0525   & 0.0692   \\
200K      & 0.0350   & 0.0476   & 0.0434   & 0.0568   & 0.0628   \\
300K      & -0.0235  & -0.0324  & -0.0316  & -0.0044  & -0.0116  \\
400K      & -0.1236  & -0.1056  & -0.1133  & 0.0232   & 0.0130   \\
500K      & 0.0346   & -0.0368  & -0.0246  & -0.0063  & -0.0409  \\
600K      & -0.1185  & -0.0546  & 0.0645   & -0.0039  & 0.0372   \\
4M        & -0.2065  & -0.1279  & -0.1928  & -0.3621  & -0.4942 
\end{tabular}
  \captionof{table}{Detailed cosine similarity results of the $8^{\text{th}}$ block in SiT-XL/2 at $t\leq0.10$.}
  \label{fig:app_grad_sim_tab}
  \vspace{1em}
\end{minipage}
\newline
We first observe a relatively high cosine similarity, representing an acute angle between gradients of the two objectives.
However, the similarity shows a decreasing trend as the training progresses, and the angle becomes nearly orthogonal at the intermediate stage (around 400K iteration).
Furthermore, we find that the similarity becomes obviously negative at the final training stage, such as at 4M iteration, indicating that there might be some potential conflict between REPA and diffusion loss.

In addition to training iterations, we also find a feature alignment gap over different diffusion timesteps: 
As reported in \citep{yu2025repa}, a well-trained DiT~\citep{Peebles2022DiT} or SiT exhibits a higher feature alignment at the intermediate diffusion timesteps, while the alignment is notably weaker at those closer to the data distribution, i.e., nearby the sampling results, such as $t=0.1$ for SiT.
We observe a similar trend in our gradient similarity comparison.
According to diffusion sampling properties, the initial steps starting from noise mainly contribute to global fidelity, namely the basic outline of images, while the steps closer to the data are to refine microscopic details such as textures \citep{DBLP:conf/icml/KimSSKM22}.
We hypothesize that the diffusion transformer eventually needs to refine its own representations for detail generation beyond learning directly from external features.

\begin{table}[h]
\tablestyle{7pt}{1.3}
  \centering
  \begin{tabular}{lccccccc}
iteration & t = 0.02 & t = 0.05 & t = 0.07 & t = 0.1 & t = 0.2 & t = 0.5 & t = 0.9\\ \midrule[1.2pt]
100K      & -0.0138   & -0.0131   & -0.0068   & 0.0129   & 0.0488  & 0.0541 & -0.0093\\
200K      & -0.0423   & -0.0674   & -0.0719   & -0.0491   & 0.0068 & 0.0801 & 0.0099\\
250K      & -0.0323  & -0.0597  & -0.0598  & -0.0599  & -0.0264  & 0.0354 & 0.0419\\
260K      & -0.0232  & -0.0331  & -0.0243  & -0.0034   & 0.0436   & 0.0729 & 0.0065\\
270K      &  0.0029  &  0.0152  & 0.0113  & 0.0097  & 0.0419  & 0.0554 & 0.0233\\
280K      & -0.0263  & -0.0131  & -0.0031   & 0.0011  & 0.0217   & 0.0455 & -0.0176\\
290K      & -0.0524  &  0.0199  & 0.0308  & 0.0532  & 0.0832 & 0.0550 & 0.0111
\end{tabular}
  \captionof{table}{Detailed gradient cosine similarity results between holistic alignment and denoising objectives on the $8^{\text{th}}$ block of SiT-XL/2 at different training iterations.}
  \label{fig:app_grad_sim_tab_}
\end{table}

For our method, HASTE, we also examine the gradient cosine similarity between holistic alignment and denoising. The similarity trend serves as a kind of reference for our termination strategy.

\newpage
\subsection{Detailed Quantitative Results}
We provide detailed evaluation results of HASTE on different SiT models in Table~\ref{tab:app_sit_results}. 
All results are reported with the SDE Euler-Maruyama sampler ($\text{NFEs} = 250$) and without classifier-free guidance.

\begin{table}[htbp]
\tablestyle{6pt}{1.3}
\centering
\begin{tabular}{lccccccc}
model           & \#params & iteration & FID$\downarrow$~\citep{10.5555/3295222.3295408} & sFID$\downarrow$~\citep{pmlr-v139-nash21a} & IS$\uparrow$~\citep{IS}    & Prec.$\uparrow$~\citep{precandrec} & Rec.$\uparrow$~\citep{precandrec} \\ \midrule[1.2pt]

SiT-B/2~\citep{ma2024sitexploringflowdiffusionbased}        & 130M    & 400K      & 33.0            & 6.46             & 43.7           & 0.53           & 0.63          \\
\rowcolor{gray!8}
\textbf{+HASTE} & 130M    & 100K      & 39.9            & 7.16             & 35.8           & 0.52           & 0.61          \\
\rowcolor{gray!8}
\textbf{+HASTE} & 130M    & 200K      & 25.7            & 6.66             & 57.0           & 0.59           & 0.62          \\
\rowcolor{gray!8}
\textbf{+HASTE} & 130M    & 400K      & \textbf{19.6}   & \textbf{6.38}    & \textbf{73.0}  & \textbf{0.62}  & \textbf{0.64} \\ \hline
 
 SiT-L/2~\citep{ma2024sitexploringflowdiffusionbased} & 458M & 400K & 18.8 & 5.29 & 72.0 & 0.64 & 0.64 \\
\rowcolor{gray!8}
\textbf{+HASTE} & 458M    & 100K      & 19.6            & 5.70             & 67.9           & 0.64           & 0.63          \\
\rowcolor{gray!8}
\textbf{+HASTE} & 458M    & 200K      & 12.1            & 5.28             & 96.1           & 0.68           & 0.64          \\
\rowcolor{gray!8}
\textbf{+HASTE} & 458M    & 400K      & \textbf{8.9}    & \textbf{5.18}    & \textbf{118.9} & \textbf{0.69}  & \textbf{0.66} \\ \hline
 
SiT-XL/2~\citep{ma2024sitexploringflowdiffusionbased}        & 675M    & 7M        & 8.6             & 6.32             & 131.7          & 0.68           & 0.67          \\
\rowcolor{gray!8}
\textbf{+HASTE} & 675M    & 100K      & 15.9            & 5.64             & 78.1           & 0.67           & 0.62          \\
\rowcolor{gray!8}
\textbf{+HASTE} & 675M    & 200K      & 9.9             & 5.04             & 108.8          & 0.69           & 0.64          \\
\rowcolor{gray!8}
\textbf{+HASTE} & 675M    & 250K      & 8.4             & 4.90             & 119.6          & 0.70           & \textbf{0.65} \\
\rowcolor{gray!8}
\textbf{+HASTE} & 675M    & 400K      & 7.3             & 5.05             & 128.7          & 0.72           & 0.64          \\
\rowcolor{gray!8}
\textbf{+HASTE} & 675M    & 500K      & \textbf{5.3}    & \textbf{4.72}    & \textbf{148.5} & \textbf{0.73}  & \textbf{0.65}
\end{tabular}

\caption{Additional evaluation results on ImageNet 256 $\times$ 256. $\uparrow$ and $\downarrow$ denote higher and lower values are better, respectively. \textbf{Bold font} denotes the best performance.}
\label{tab:app_sit_results}
\end{table}

Additionally, we provide the results of SiT-XL/2+HASTE with different classifier-free guidance~\citep{ho2022classifierfreediffusionguidance} scales and intervals~\citep{Kynkaanniemi2024}.

\begin{table}[htbp]
\tablestyle{6pt}{1.3}
\centering

\begin{tabular}{lccccccccc}
model           & \#params & iteration & interval  & CFG scale & FID$\downarrow$ & sFID$\downarrow$ & IS$\uparrow$    & Prec.$\uparrow$ & Rec.$\uparrow$ \\ \midrule[1.2pt]
SiT-XL/2        & 675M & 7M   & [0, 1]   & 1.50  & 2.06 & 4.50 & 270.3 & \textbf{0.82} & 0.59 \\
\rowcolor{gray!8} 
\textbf{+HASTE} & 675M & 500K & [0, 1]   & 1.25  & 2.18 & 4.67 & 240.4 & 0.81          & 0.60 \\
\rowcolor{gray!8} 
\textbf{+HASTE} & 675M & 500K & [0, 0.7] & 1.50  & 1.80 & 4.58 & 252.1 & 0.80          & 0.61 \\
\rowcolor{gray!8} 
\textbf{+HASTE} & 675M & 500K & [0, 0.6] & 1.825 & 1.74 & 4.74 & 268.7 & 0.80          & 0.62 \\
\rowcolor{gray!8} 
\textbf{+HASTE} & 675M & 2M   & [0, 0.7] & 1.7   & 1.45 & 4.55 & 297.3 & 0.80          & 0.64 \\
\rowcolor{gray!8} 
\textbf{+HASTE} & 675M & 2M   & [0, 0.7] & 1.65  & 1.44 & 4.56 & 289.4 & 0.79          & 0.64 \\
\rowcolor{gray!8} 
\textbf{+HASTE} & 675M & 2M   & [0, 0.7] & 1.675 & 1.44 & 4.55 & 293.7 & 0.80          & 0.64 \\
\rowcolor{gray!8} 
\textbf{+HASTE} & 675M & 2.5M & [0, 0.7] & 1.7   & 1.43 & 4.56 & 298.8 & 0.80          & 0.64 \\
\rowcolor{gray!8} 
\textbf{+HASTE} & 675M & 2.5M & [0, 0.7] & 1.65  & 1.43 & 4.57 & 290.7 & 0.80          & 0.64 \\
\rowcolor{gray!8} 
\textbf{+HASTE} & 675M    & 2.5M      & [0, 0.72] & 1.65      & \textbf{1.42}   & \textbf{4.49}    & \textbf{299.5} & 0.80           & \textbf{0.65}
\end{tabular}

\caption{Evaluation results on ImageNet 256 $\times$ 256 with different classifier-free guidance settings.}
\label{tab:app_cfg_results}
\end{table}

\newpage
\section{Additional Implementation Details.}

\begin{table}[htbp]
\tablestyle{6.5pt}{1.3}
\centering

\begin{tabular}{lcccc}
                      & SiT-B           & SiT-L           & SiT-XL         & DiT-XL                  \\ \midrule[1.2pt]
\textbf{Architecture} &                 &                 &                &                         \\
input dim.            & 32$\times$32$\times$4 & 32$\times$32$\times$4 & 32$\times$32$\times$4 & 32$\times$32$\times$4 \\
num. layers           & 12              & 24              & 28             & 28                      \\
hidden dim.           & 768             & 1024            & 1152           & 1152                    \\
num. heads            & 12              & 16              & 16             & 16                      \\ \hline
\textbf{HASTE}        &                 &                 &                &                         \\
$\lambda_R$           & 0.5             & 0.5             & 0.5            & 0.5                     \\
$\lambda_A$           & 0.5             & 0.5             & 0.5            & 0.5                     \\
alignment depth       & 5               & 8               & 8              & 8                       \\
student layers        & [2, 3, 4]       & [4, 5, 6, 7]    & [4, 5, 6, 7]   & [4, 5, 6, 7]            \\
teacher model         & DINOv2-B~\citep{oquab2024dinov}        & DINOv2-B~\citep{oquab2024dinov}        & DINOv2-B~\citep{oquab2024dinov}       & DINOv2-B~\citep{oquab2024dinov}                \\
teacher layers        & [7, 9, 11]      & [8, 9, 10, 11]  & [8, 9, 10, 11] & [8, 9, 10, 11]          \\
termination iter.     & 100 K           & 250 K           & 250 K          & 250 K                   \\
alignment heads       & 0-11            & 0-11            & 0-11           & 0-11                    \\ \hline
\textbf{Optimization} & \multicolumn{1}{l}{}  & \multicolumn{1}{l}{}  & \multicolumn{1}{l}{}  & \multicolumn{1}{l}{}  \\
batch size            & 256             & 256             & 256            & 256                     \\
optimizer             & AdamW~\citep{kingma2017adammethodstochasticoptimization, loshchilov2019decoupledweightdecayregularization}           & AdamW~\citep{kingma2017adammethodstochasticoptimization, loshchilov2019decoupledweightdecayregularization}            & AdamW~\citep{kingma2017adammethodstochasticoptimization, loshchilov2019decoupledweightdecayregularization}           & AdamW~\citep{kingma2017adammethodstochasticoptimization, loshchilov2019decoupledweightdecayregularization}                    \\
lr                    & 0.0001          & 0.0001          & 0.0001         & 0.0001                  \\
($\beta_1$, $\beta_2$)    & (0.9, 0.999)    & (0.9, 0.999)    & (0.9, 0.999)   & (0.9, 0.999)            \\
weight decay & 0 & 0 & 0 & 0 \\\hline
\textbf{Diffusion}    & \multicolumn{1}{l}{}  & \multicolumn{1}{l}{}  & \multicolumn{1}{l}{}  & \multicolumn{1}{l}{}  \\
objective             & linear interpolants & linear interpolants & linear interpolants & improved DDPM           \\
prediction            & velocity             & velocity             & velocity            & noise and variance \\
sampler               & Euler-Maruyama  & Euler-Maruyama  & Euler-Maruyama & Euler-Maruyama          \\
sampling steps        & 250             & 250             & 250            & 250                    
\end{tabular}

\caption{Detailed training settings.}
\label{tab:app_imp_detail}
\end{table}

\paragraph{Further implementation details.} 
For XL and L-sized models, we set the feature alignment depth to 8 following REPA, and extract the attention maps from layer [4, 5, 6, 7] (counting from 0) of diffusion transformers, to align with those from layer [8, 9, 10, 11] of DINOv2-B. 
According to \citep{li2024on}, the performance almost saturates when transferring 12 out of 16 heads, and the student can also develop its own attention patterns for unused heads. 
Specifically, since the number of heads for DINOv2-B layer is only 12, we conduct attention alignment partially over the first 12 heads of diffusion transformer layer. 
For B-sized models, the feature alignment depth is adjusted to 5, and we extract the attention maps from layer [2, 3, 4] to align with those from layer [7, 9, 11] of DINOv2-B. 

We enable mixed-precision (fp16) for efficient training. For data pre-processing, we leverage the protocols provided in EDM2~\citep{Karras2024edm2} to pre-compute latent vectors from images with stable diffusion VAE~\citep{LDM}. Specifically, we use \texttt{stabilityai/sd-vae-ft-ema} decoder to translate generated latent vectors into images. Following REPA~\citep{yu2025repa}, we also use three-layer MLP with SiLU activations~\citep{SiLu2018} as the projector of hidden states. For MM-DiT, we use CLIP~\citep{radford2021learning} text model to encode captions.

\newpage
\section{Additional Visualizations}

\begin{figure}[ht]
    \centering
    \includegraphics[width=1\linewidth]{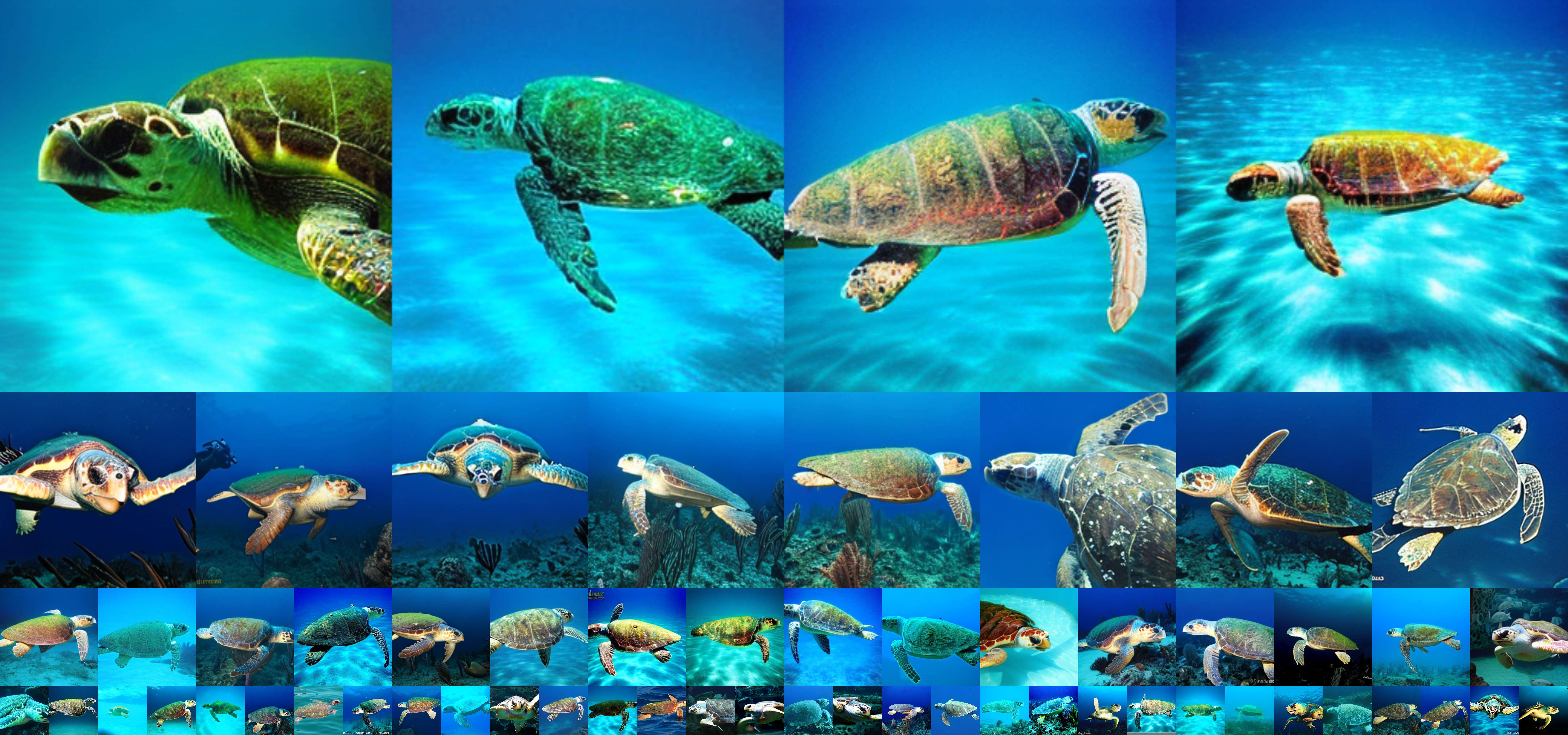}
    \caption{Uncurated generation results of SiT-XL/2+HASTE. We use classifier-free guidance
 with~$w = 4.0$. Class label = ``loggerhead sea turtle'' (33).}
    \label{fig:vis33}
\end{figure}

\begin{figure}[ht]
    \centering
    \includegraphics[width=1\linewidth]{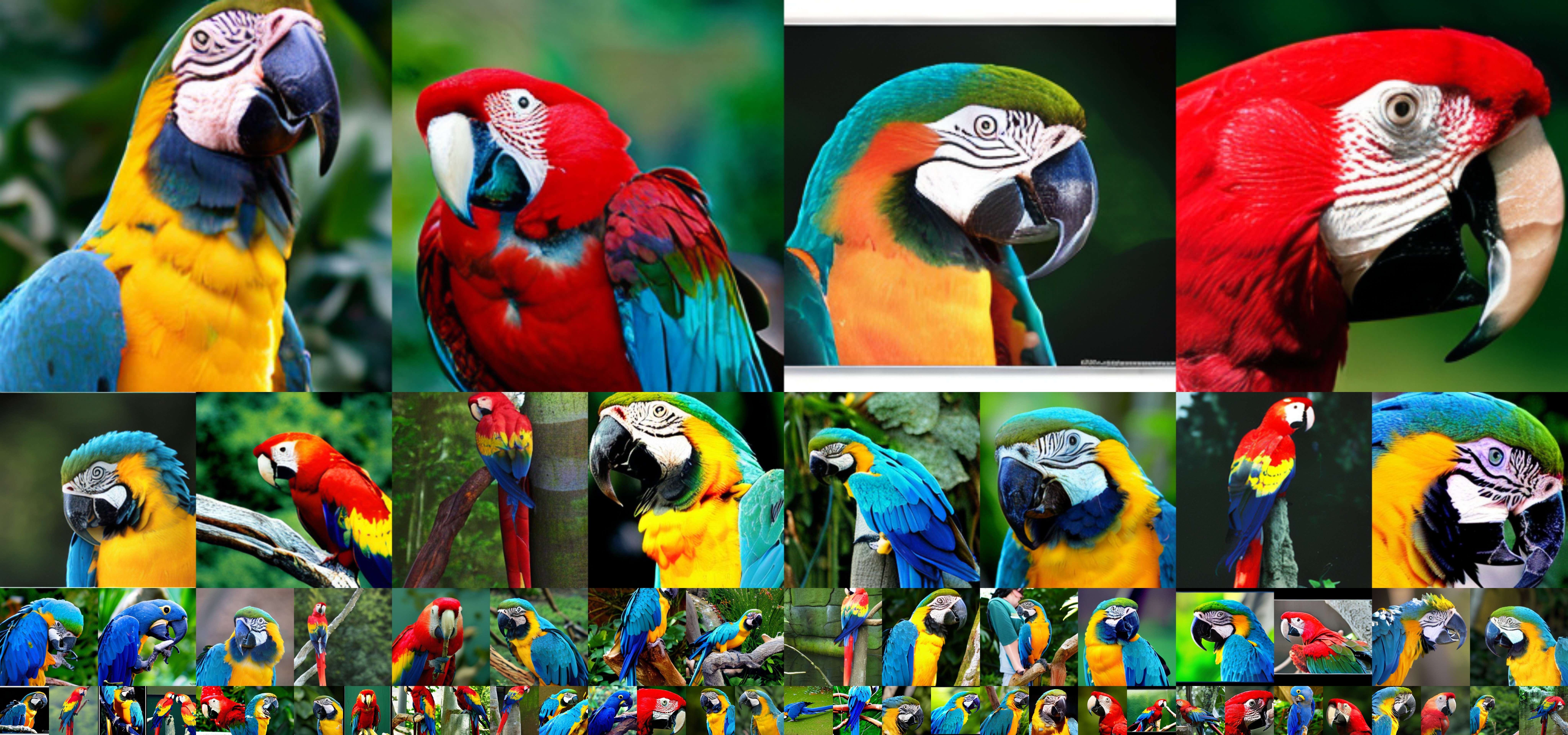}
    \caption{Uncurated generation results of SiT-XL/2+HASTE. We use classifier-free guidance
 with~$w = 4.0$.  Class label = ``macaw'' (88).}
    \label{fig:vis88}
\end{figure}

\newpage

\begin{figure}[ht]
    \centering
    \includegraphics[width=1\linewidth]{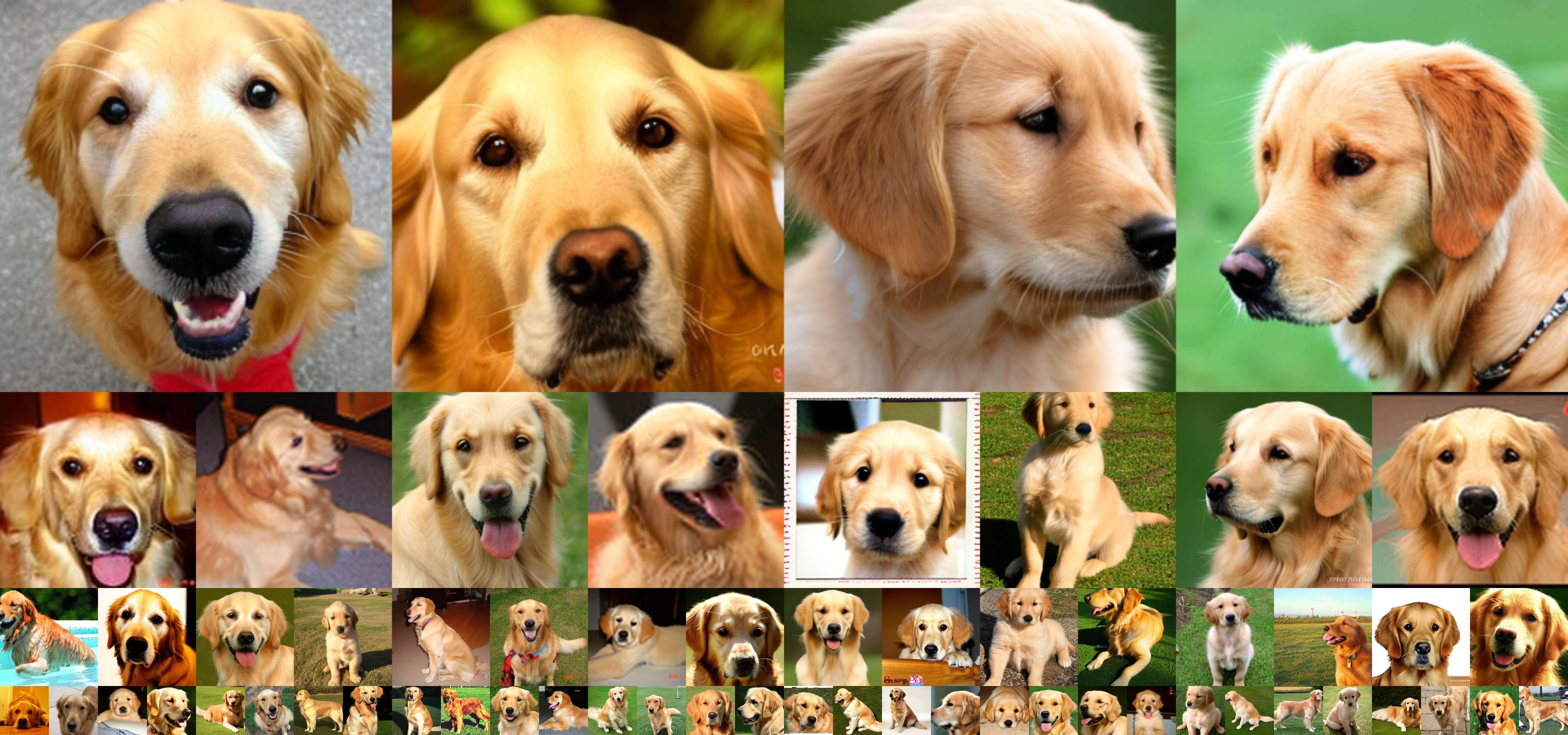}
    \caption{Uncurated generation results of SiT-XL/2+HASTE. We use classifier-free guidance
 with~$w = 4.0$. Class label = ``golden retriever'' (207).}
    \label{fig:vis207}
\end{figure}

\begin{figure}[ht]
    \centering
    \includegraphics[width=1\linewidth]{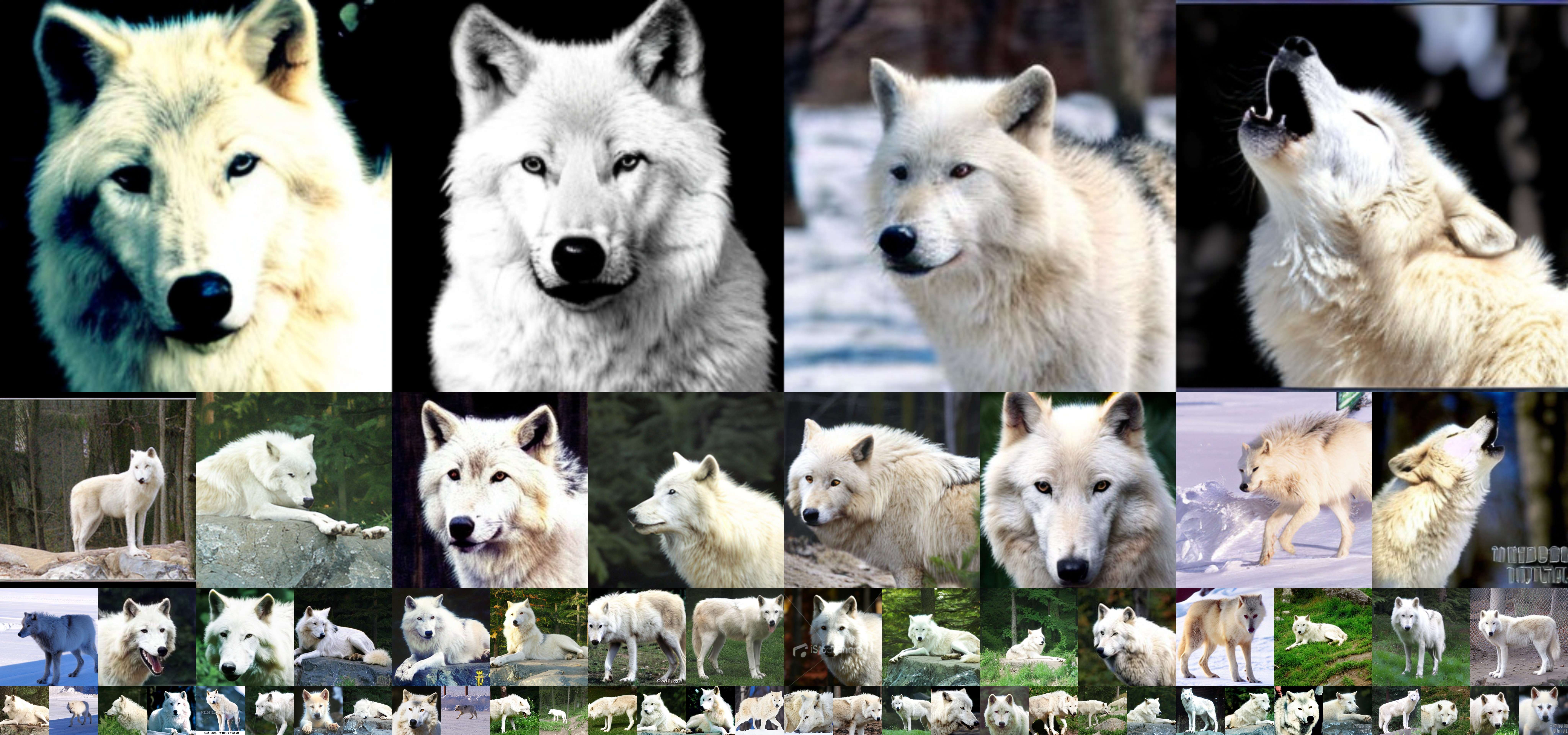}
    \caption{Uncurated generation results of SiT-XL/2+HASTE. We use classifier-free guidance
 with~$w = 4.0$.  Class label = ``arctic wolf'' (270).}
    \label{fig:vis270}
\end{figure}

\newpage

\begin{figure}[ht]
    \centering
    \includegraphics[width=1\linewidth]{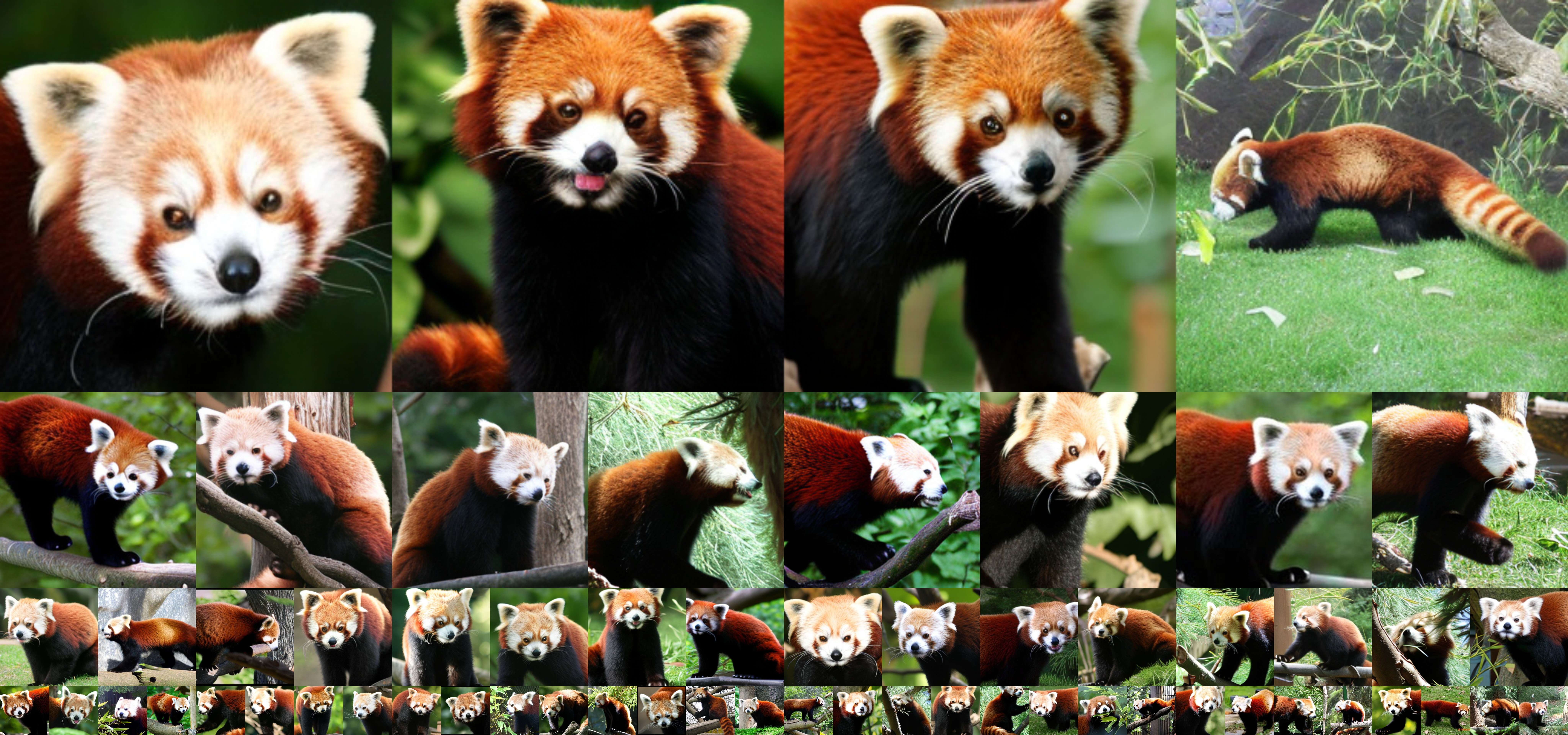}
    \caption{Uncurated generation results of SiT-XL/2+HASTE. We use classifier-free guidance
 with~$w = 4.0$. Class label = ``red panda'' (387).}
    \label{fig:vis387}
\end{figure}

\begin{figure}[ht]
    \centering
    \includegraphics[width=1\linewidth]{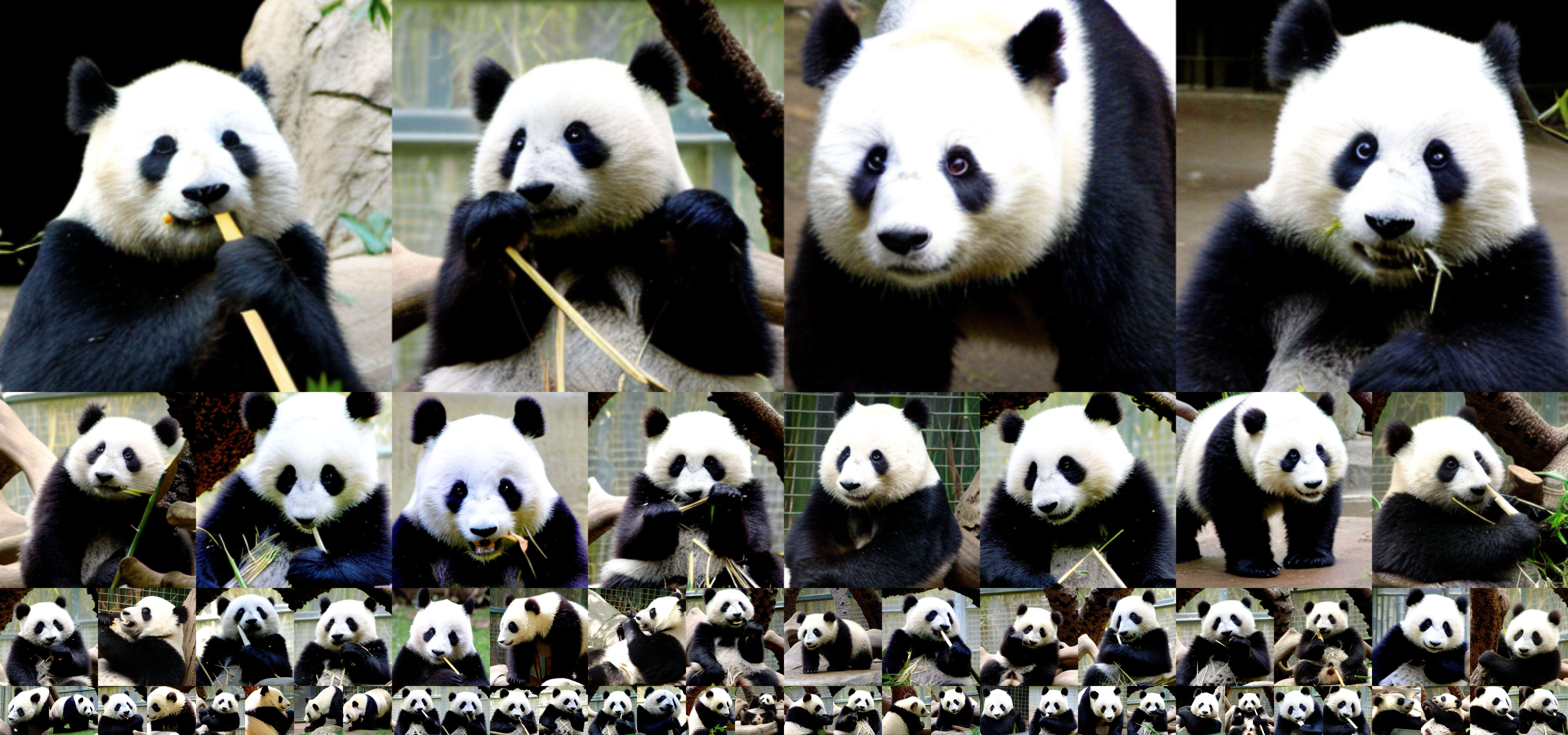}
    \caption{Uncurated generation results of SiT-XL/2+HASTE. We use classifier-free guidance
 with~$w = 4.0$.  Class label = ``panda'' (388).}
    \label{fig:vis388}
\end{figure}

\newpage

\begin{figure}[ht]
    \centering
    \includegraphics[width=1\linewidth]{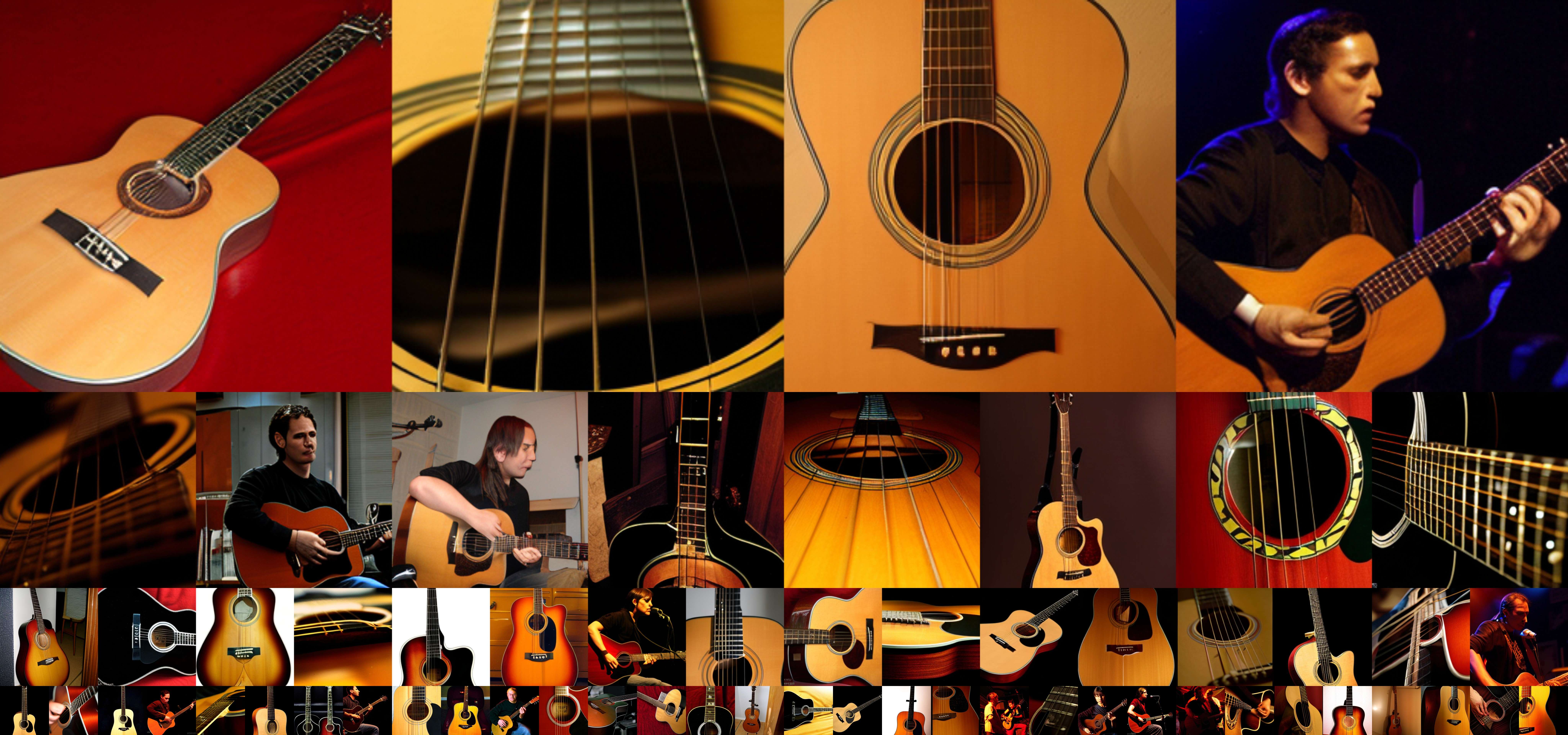}
    \caption{Uncurated generation results of SiT-XL/2+HASTE. We use classifier-free guidance
 with~$w = 4.0$. Class label = ``acoustic guitar'' (402).}
    \label{fig:vis402}
\end{figure}

\begin{figure}[ht]
    \centering
    \includegraphics[width=1\linewidth]{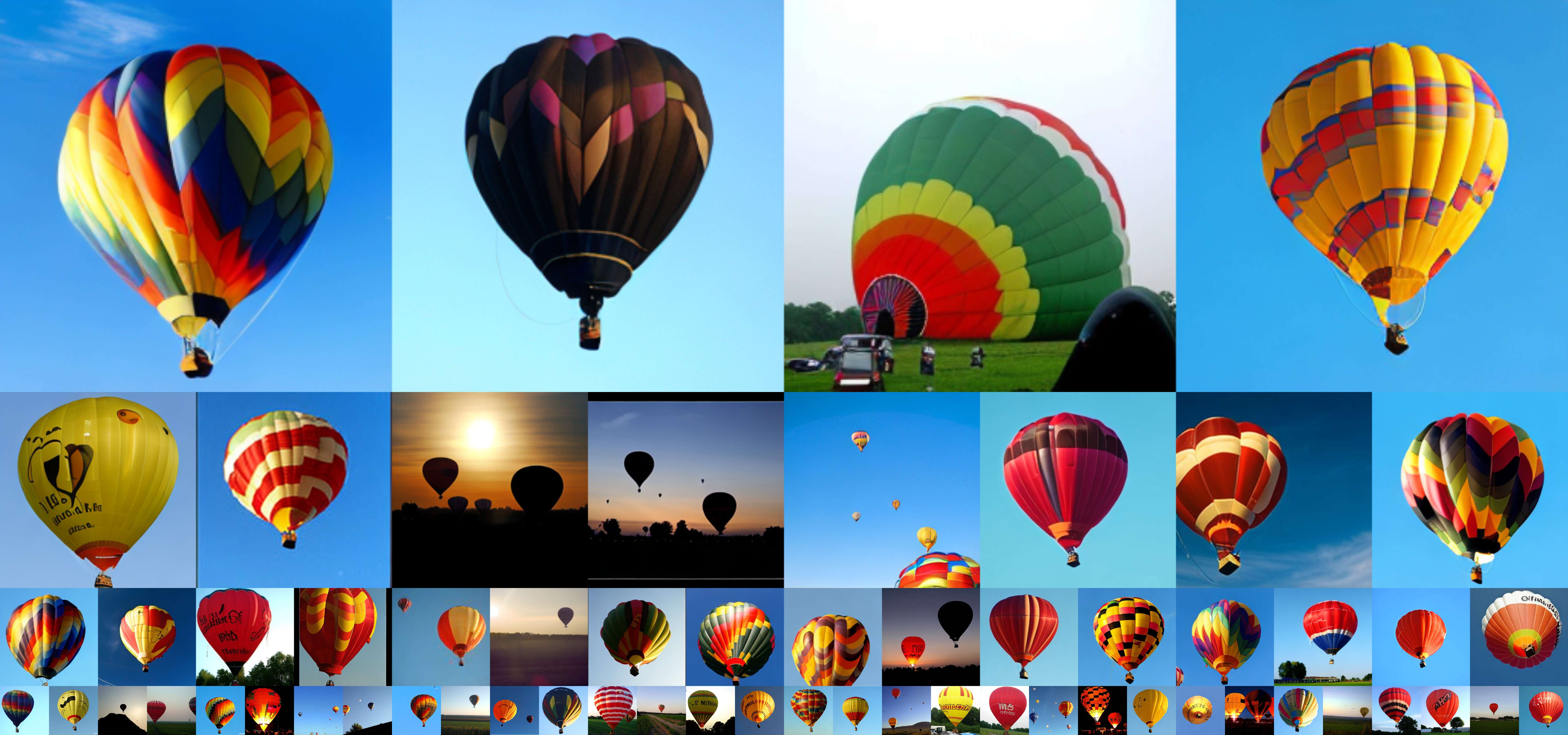}
    \caption{Uncurated generation results of SiT-XL/2+HASTE. We use classifier-free guidance
 with~$w = 4.0$.  Class label = ``balloon'' (417).}
    \label{fig:vis417}
\end{figure}

\newpage

\begin{figure}[ht]
    \centering
    \includegraphics[width=1\linewidth]{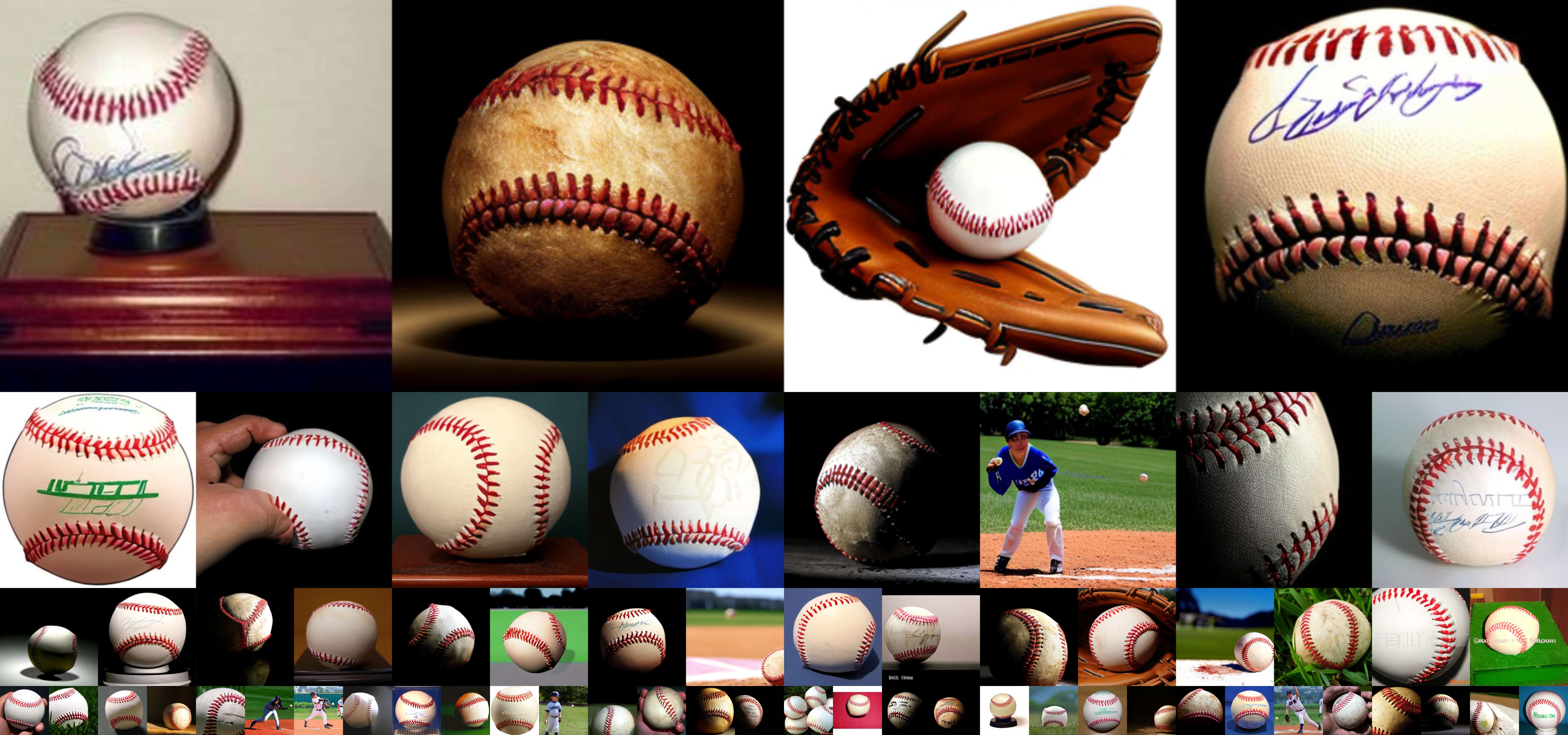}
    \caption{Uncurated generation results of SiT-XL/2+HASTE. We use classifier-free guidance
 with~$w = 4.0$. Class label = ``baseball'' (429).}
    \label{fig:vis429}
\end{figure}

\begin{figure}[ht]
    \centering
    \includegraphics[width=1\linewidth]{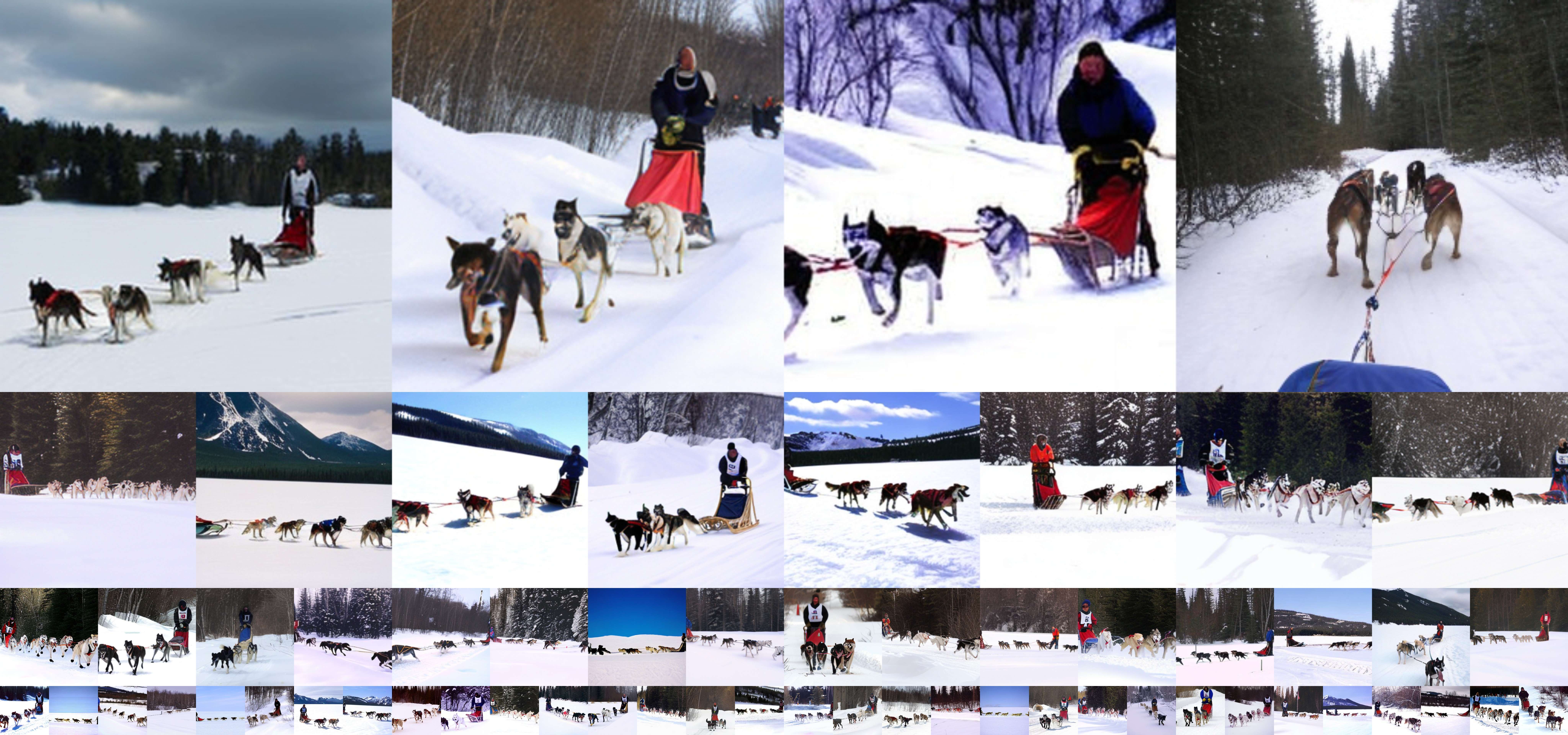}
    \caption{Uncurated generation results of SiT-XL/2+HASTE. We use classifier-free guidance
 with~$w = 4.0$.  Class label = ``dog sled'' (537).}
    \label{fig:vis537}
\end{figure}

\newpage

\begin{figure}[ht]
    \centering
    \includegraphics[width=1\linewidth]{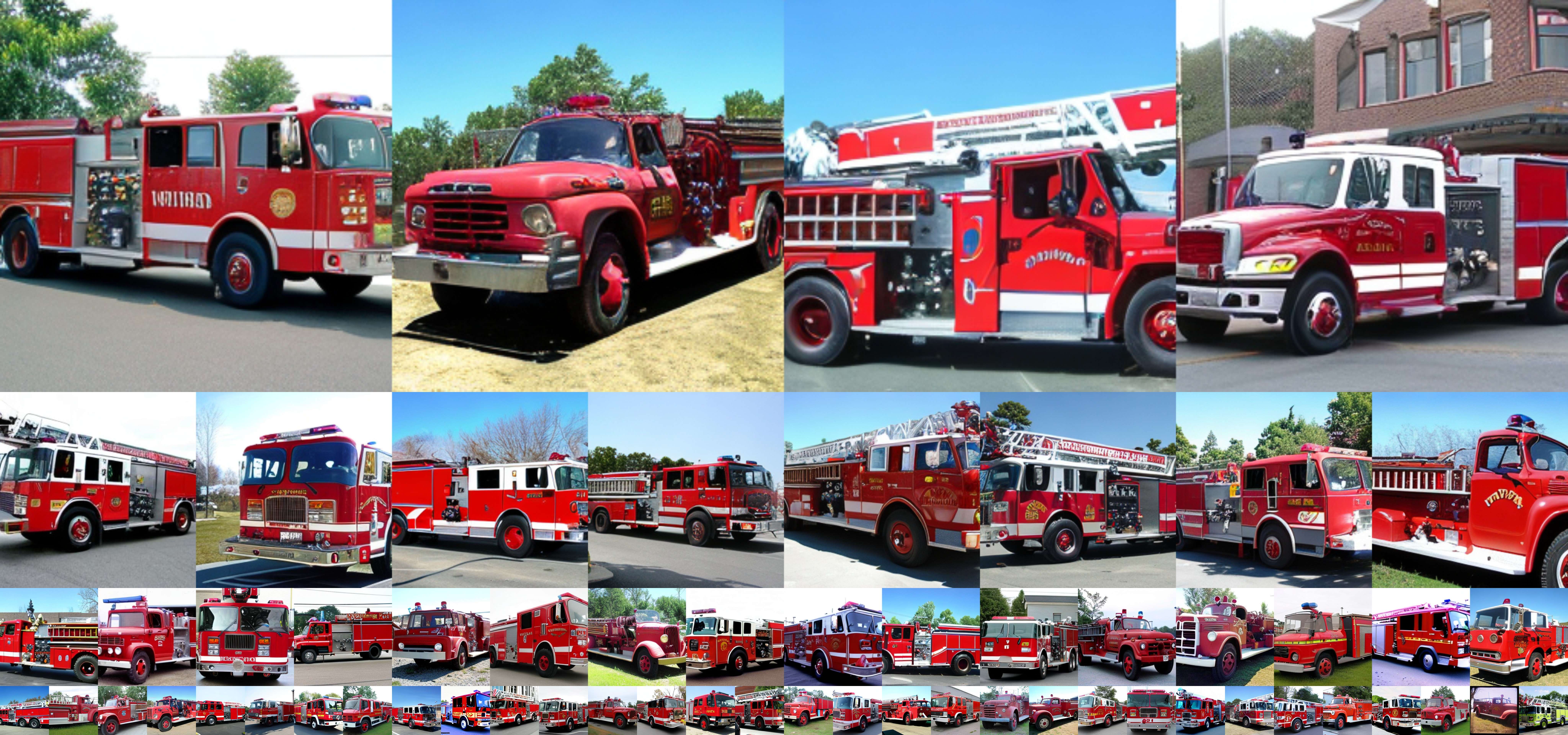}
    \caption{Uncurated generation results of SiT-XL/2+HASTE. We use classifier-free guidance
 with~$w = 4.0$. Class label = ``fire truck'' (555).}
    \label{fig:vis555}
\end{figure}

\begin{figure}[ht]
    \centering
    \includegraphics[width=1\linewidth]{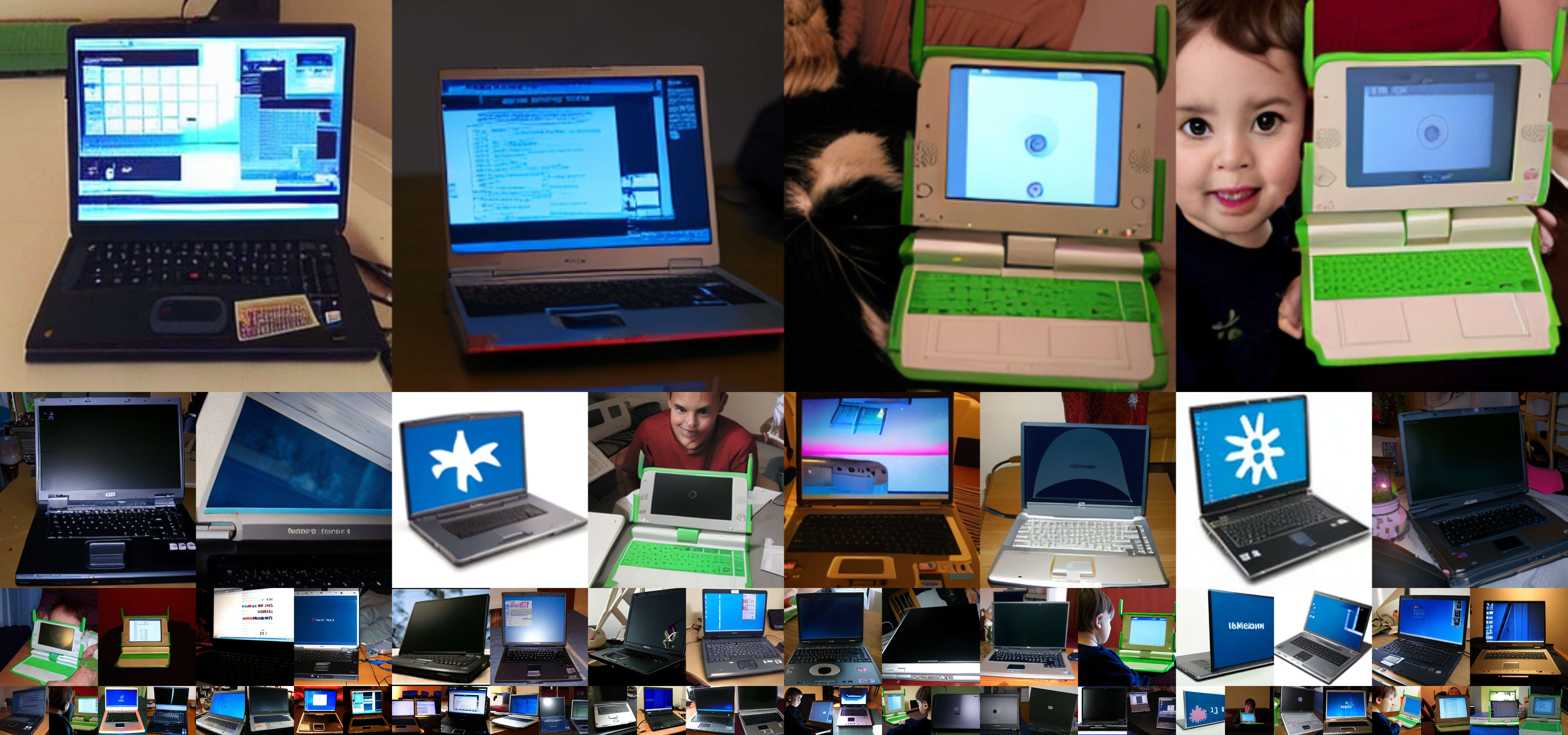}
    \caption{Uncurated generation results of SiT-XL/2+HASTE. We use classifier-free guidance
 with~$w = 4.0$.  Class label = ``laptop'' (620).}
    \label{fig:vis620}
\end{figure}

\newpage

\begin{figure}[ht]
    \centering
    \includegraphics[width=1\linewidth]{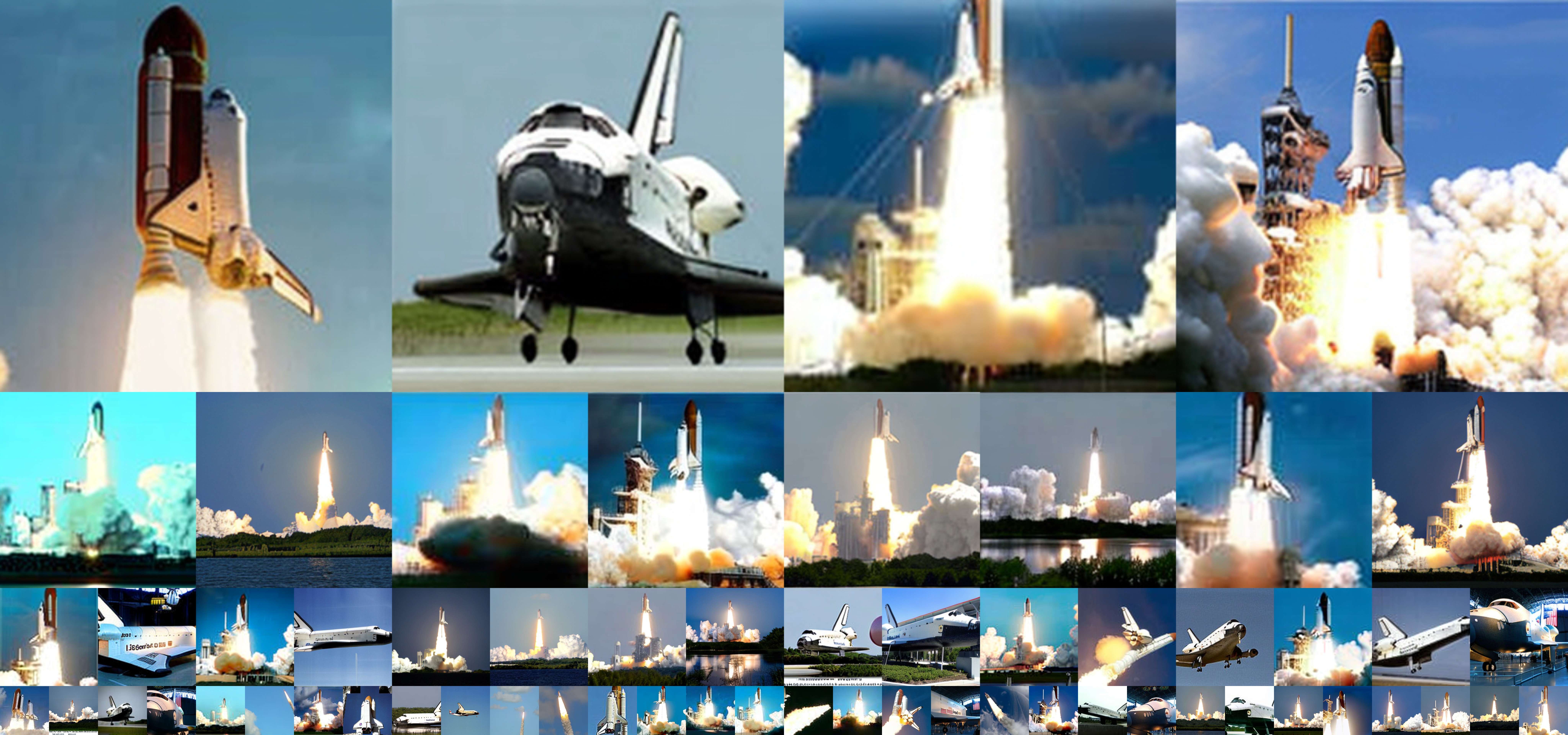}
    \caption{Uncurated generation results of SiT-XL/2+HASTE. We use classifier-free guidance
 with~$w = 4.0$. Class label = ``space shuttle'' (812).}
    \label{fig:vis812}
\end{figure}

\begin{figure}[ht]
    \centering
    \includegraphics[width=1\linewidth]{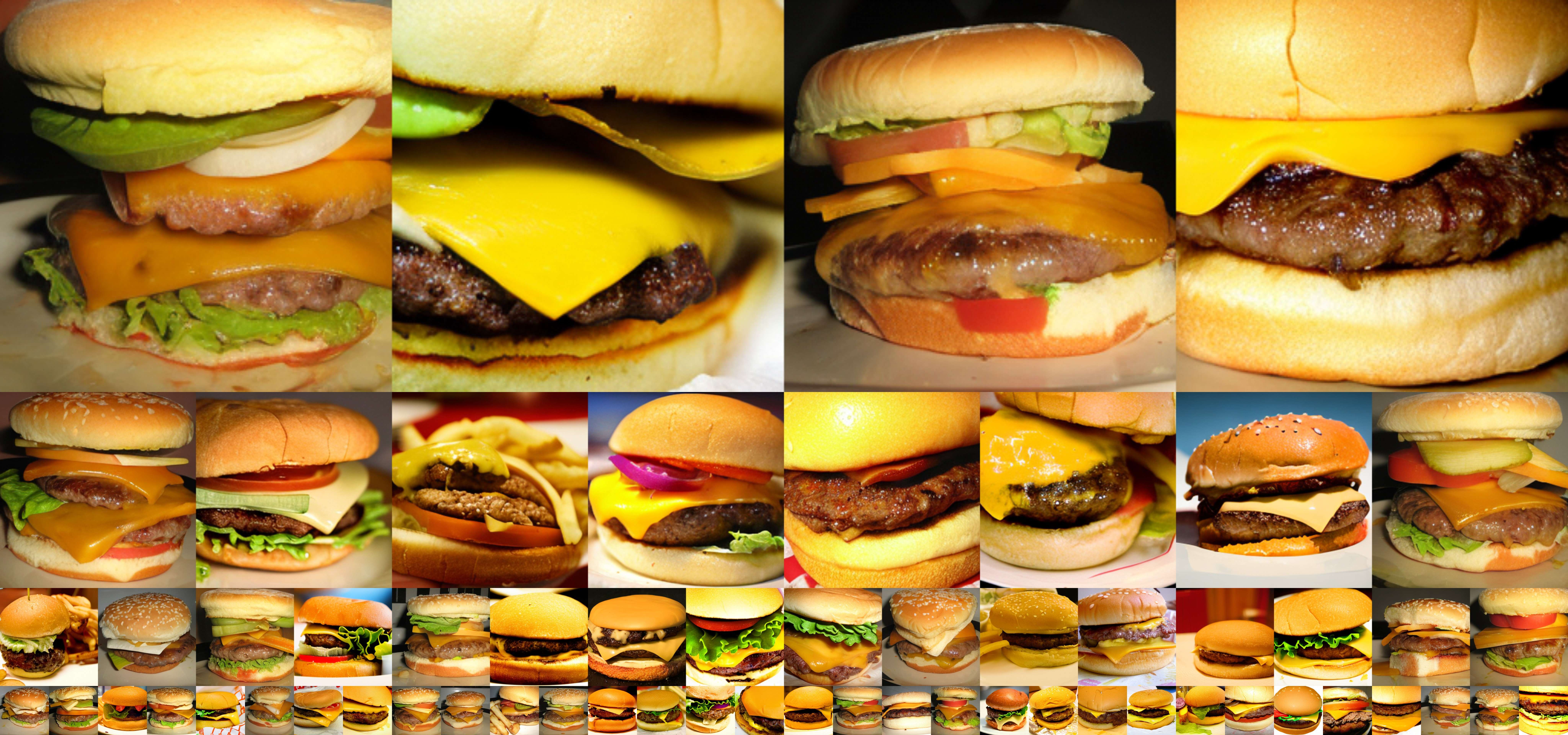}
    \caption{Uncurated generation results of SiT-XL/2+HASTE. We use classifier-free guidance
 with~$w = 4.0$.  Class label = ``cheeseburger'' (933).}
    \label{fig:vis933}
\end{figure}

\newpage

\begin{figure}[ht]
    \centering
    \includegraphics[width=1\linewidth]{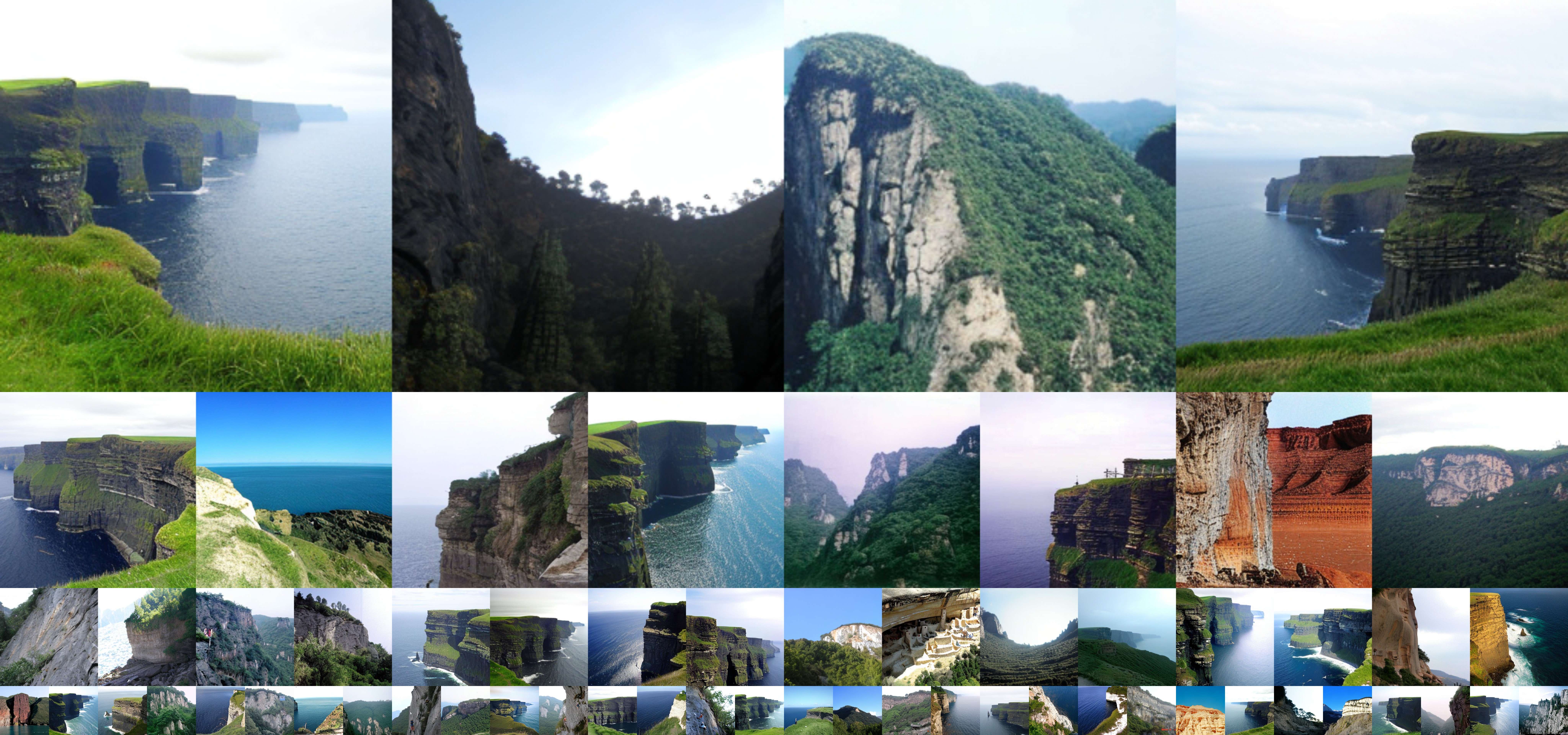}
    \caption{Uncurated generation results of SiT-XL/2+HASTE. We use classifier-free guidance
 with~$w = 4.0$. Class label = ``cliff drop-off'' (972).}
    \label{fig:vis972}
\end{figure}

\begin{figure}[ht]
    \centering
    \includegraphics[width=1\linewidth]{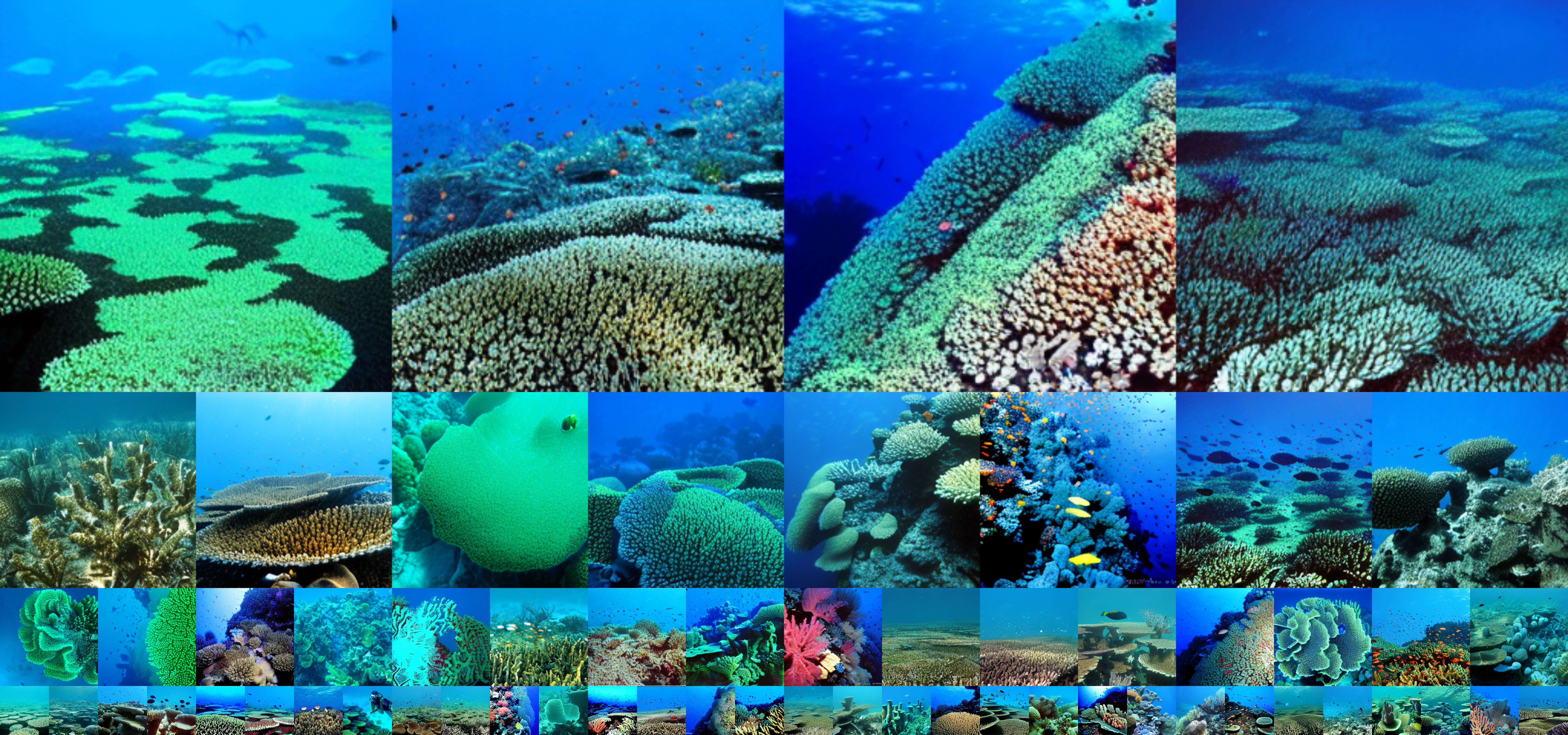}
    \caption{Uncurated generation results of SiT-XL/2+HASTE. We use classifier-free guidance
 with~$w = 4.0$.  Class label = ``coral reef'' (973).}
    \label{fig:vis973}
\end{figure}

\newpage

\begin{figure}[ht]
    \centering
    \includegraphics[width=1\linewidth]{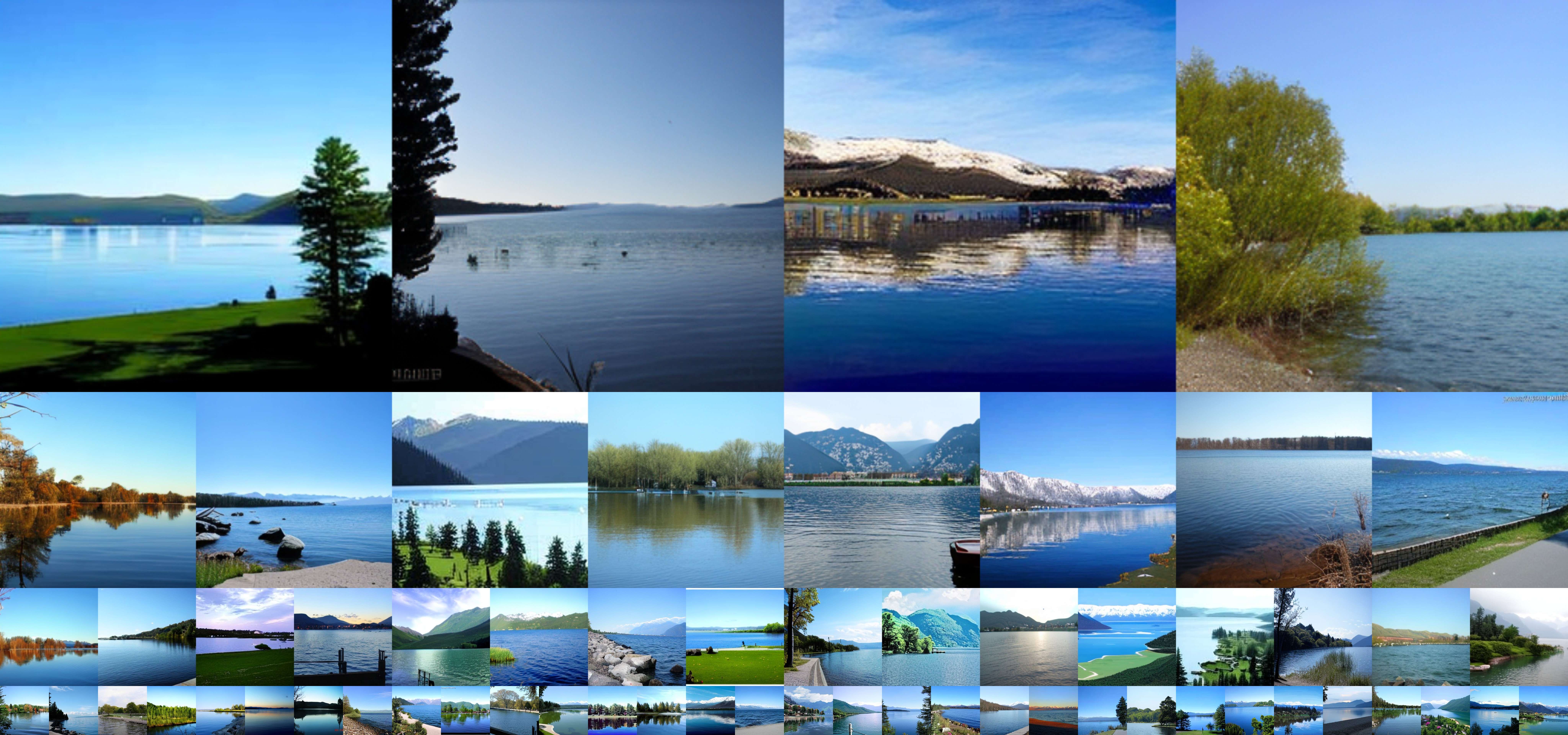}
    \caption{Uncurated generation results of SiT-XL/2+HASTE. We use classifier-free guidance
 with~$w = 4.0$. Class label = ``lake shore'' (975).}
    \label{fig:vis975}
\end{figure}

\begin{figure}[ht]
    \centering
    \includegraphics[width=1\linewidth]{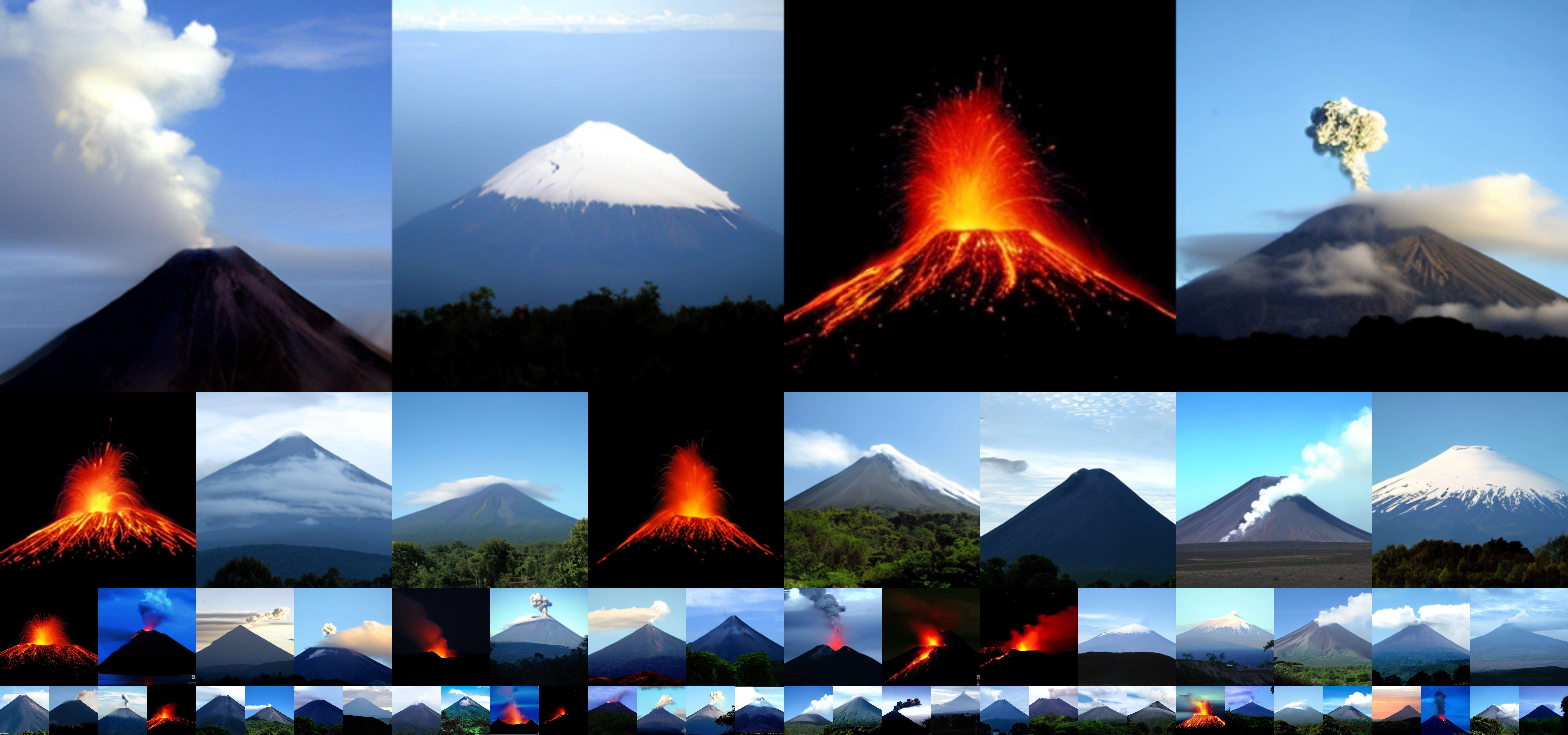}
    \caption{Uncurated generation results of SiT-XL/2+HASTE. We use classifier-free guidance
 with~$w = 4.0$.  Class label = ``volcano'' (980).}
    \label{fig:vis980}
\end{figure}

\end{document}